\documentclass{article}
\PassOptionsToPackage{square,sort,comma,numbers}{natbib}

    \usepackage[preprint]{neurips_2020}

\usepackage[utf8]{inputenc} % allow utf-8 input
\usepackage[T1]{fontenc}    % use 8-bit T1 fonts
\usepackage{hyperref}
\usepackage{url}            % simple URL typesetting
\usepackage{booktabs}       % professional-quality tables
\usepackage{amsfonts}       % blackboard math symbols
\usepackage{nicefrac}       % compact symbols for 1/2, etc.
\usepackage{microtype}      % microtypography
\usepackage{bm}
\usepackage[noend]{algpseudocode}
\usepackage{algorithm}
\usepackage{tabularx}
\usepackage{multirow}
\usepackage{graphicx}
\usepackage{amsmath,amssymb}
\usepackage{microtype}
\usepackage{subfigure}
\usepackage{booktabs} % for
\usepackage{xcolor}
\usepackage{wrapfig}

\usepackage{booktabs} 
\usepackage{colortbl} 
\usepackage[title]{appendix}
\usepackage[capitalize,nameinlink]{cleveref}

\newcommand{\bee}{\begin{eqnarray}}
\newcommand{\eee}{\end{eqnarray}}

\newlength{\widebarargwidth}
\newlength{\widebarargheight}
\newlength{\widebarargdepth}

\newcommand{\eat}[1]{}

\newcommand{\btx}{\tilde{\mathbf{x}}}
\newcommand{\bty}{\tilde{y}}

\newcommand{\bx}{\mathbf{x}}

\newcommand{\bz}{\mathbf{z}}

\usepackage{amsmath}
\usepackage{pifont}
\newcommand{\cmark}{\text{\ding{51}}}
\newcommand{\xmark}{\text{\ding{55}}}
\title{Meta-Learned Confidence for Few-shot Learning}

\author{%
  Seong Min Kye$^1$, Hae Beom Lee$^1$, Hoirin Kim$^1$, Sung Ju Hwang$^{1,2}$\\
  $^1$KAIST, $^2$AITRICS, South Korea\\
  \texttt{\{kye9165,haebeom.lee,hoirkim,sjhwang82\}@kaist.ac.kr} \\
}

\begin{document}

    \maketitle
    \begin{abstract}
Transductive inference is an effective means of tackling the data deficiency problem in few-shot learning settings. A popular transductive inference technique for few-shot metric-based approaches, is to update the prototype of each class with the mean of the most confident query examples, or confidence-weighted average of all the query samples. However, a caveat here is that the model confidence may be unreliable, which may lead to incorrect predictions. To tackle this issue, we propose to \emph{meta-learn} the confidence for each query sample, to assign optimal weights to unlabeled queries such that they improve the model's transductive inference performance on unseen tasks. We achieve this by meta-learning an input-adaptive distance metric over a task distribution under various model and data perturbations, which will enforce consistency on the model predictions under diverse uncertainties for unseen tasks. Moreover, we additionally suggest a regularization which explicitly enforces the consistency on the predictions across the different dimensions of a high-dimensional embedding vector. We validate our few-shot learning model with meta-learned confidence on four benchmark datasets, on which it largely outperforms strong recent baselines and obtains new state-of-the-art results. Further application on semi-supervised few-shot learning tasks also yields significant performance improvements over the baselines. The source code of our algorithm is available at \color{magenta}\url{https://github.com/seongmin-kye/MCT}.
\end{abstract}

    \section{Introduction}
\label{submission}
Few-shot learning, the problem of learning under data scarcity, is an important challenge in deep learning as large number of training instances may not be available in many real-world settings. While the recent advances in meta-learning made it possible to obtain impressive performance on few-shot learning tasks~\cite{hou2019cross, li2019learning,lifchitz2019dense}, it still remains challenging in cases where we are given very little information (e.g. one-shot learning). Some of the metric-based meta-learning approaches tackle this problem using \emph{transductive learning} or \emph{semi-supervised learning}, by leveraging the structure of the unlabeled instances at the inference time~\cite{hou2019cross,kim2019edge,l2ST,liu2018learning,ren2018meta}. Popular approach for these problem includes leveraging nearest neighbor graph for propagating labels~\cite{kim2019edge,liu2018learning,DPGN}, or using predicted soft or hard labels on unlabeled samples to update the class prototype~\cite{hou2019cross,ren2018meta}. However, all these transductive or semi-supervised inference approaches are fundamentally limited by the intrinsic \emph{unreliability} of the labels predicted on the unseen samples. 

\begin{figure}[t!]
	\vspace{-0.1in}
	\centering
	\hfill
	\subfigure[ProtoNets (+4.44\%)]{\includegraphics[width=4.3cm]{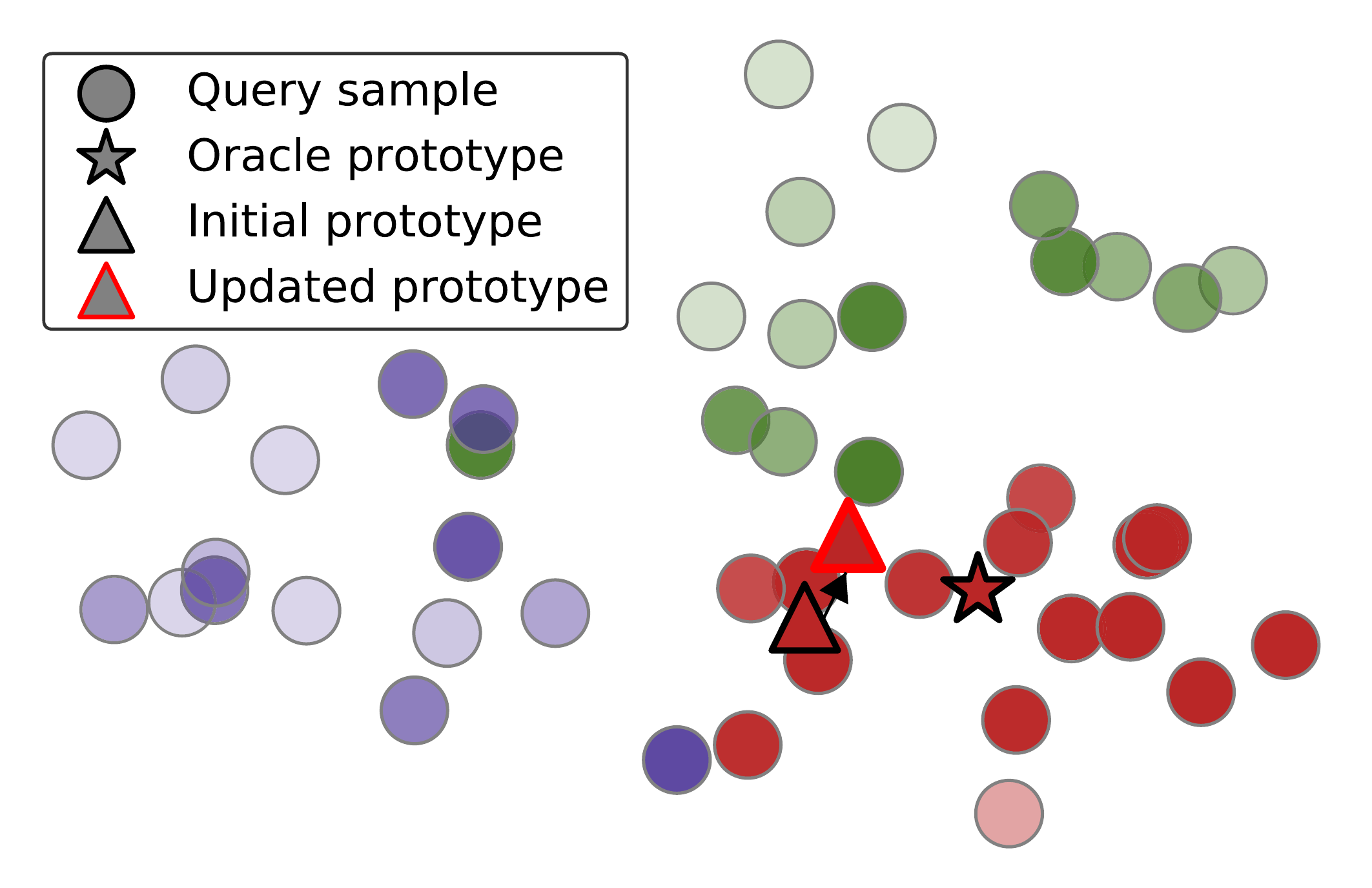}\label{fig:fig1_PN}}
    \hfill
	\subfigure[Instance-wise metric (+8.89\%) ]{\includegraphics[width=4.3cm]{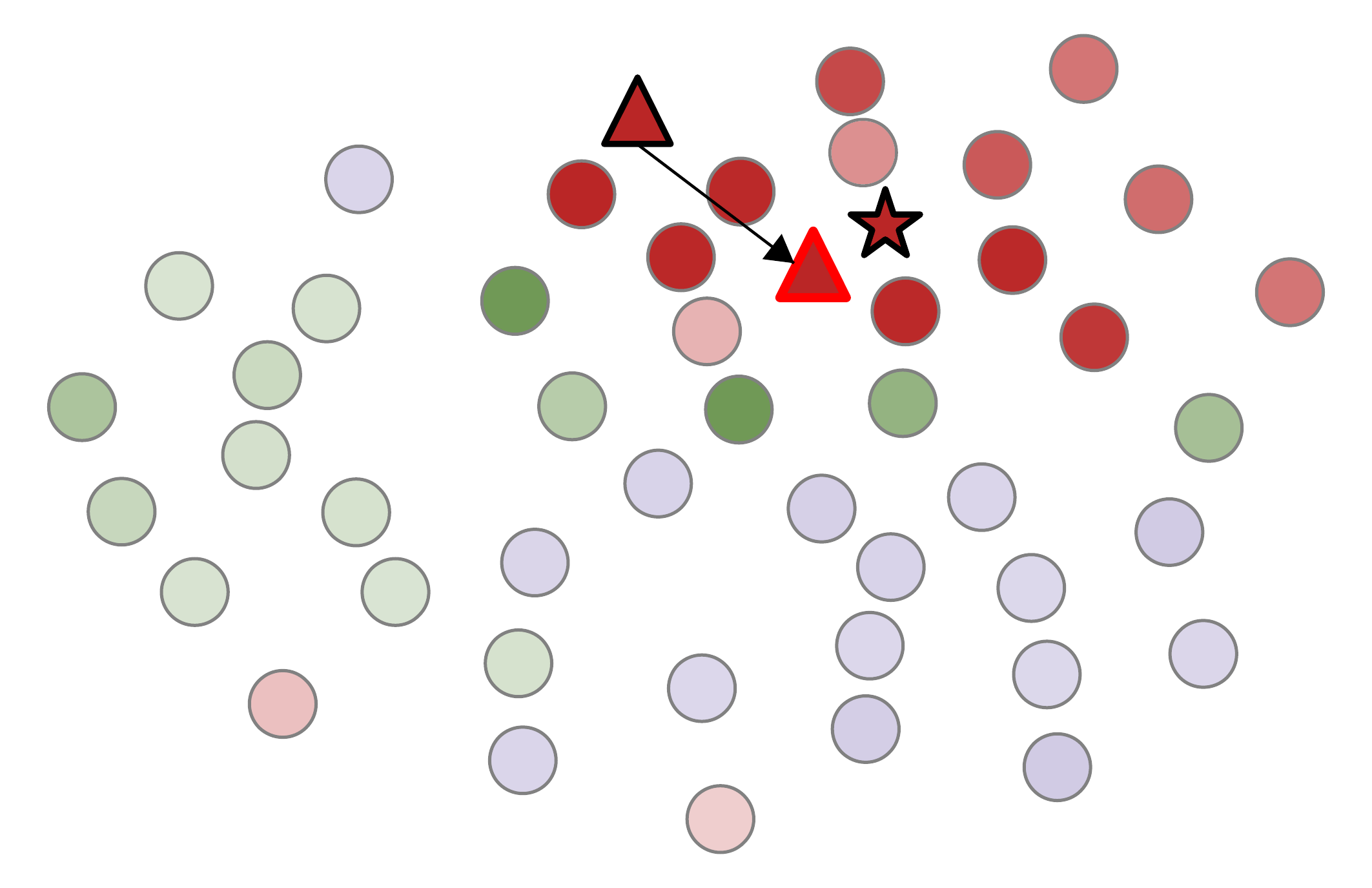}\label{fig:fig1_IMS}}
	\hfill
	\subfigure[MCT (+15.56\%)]{\includegraphics[width=4.3cm]{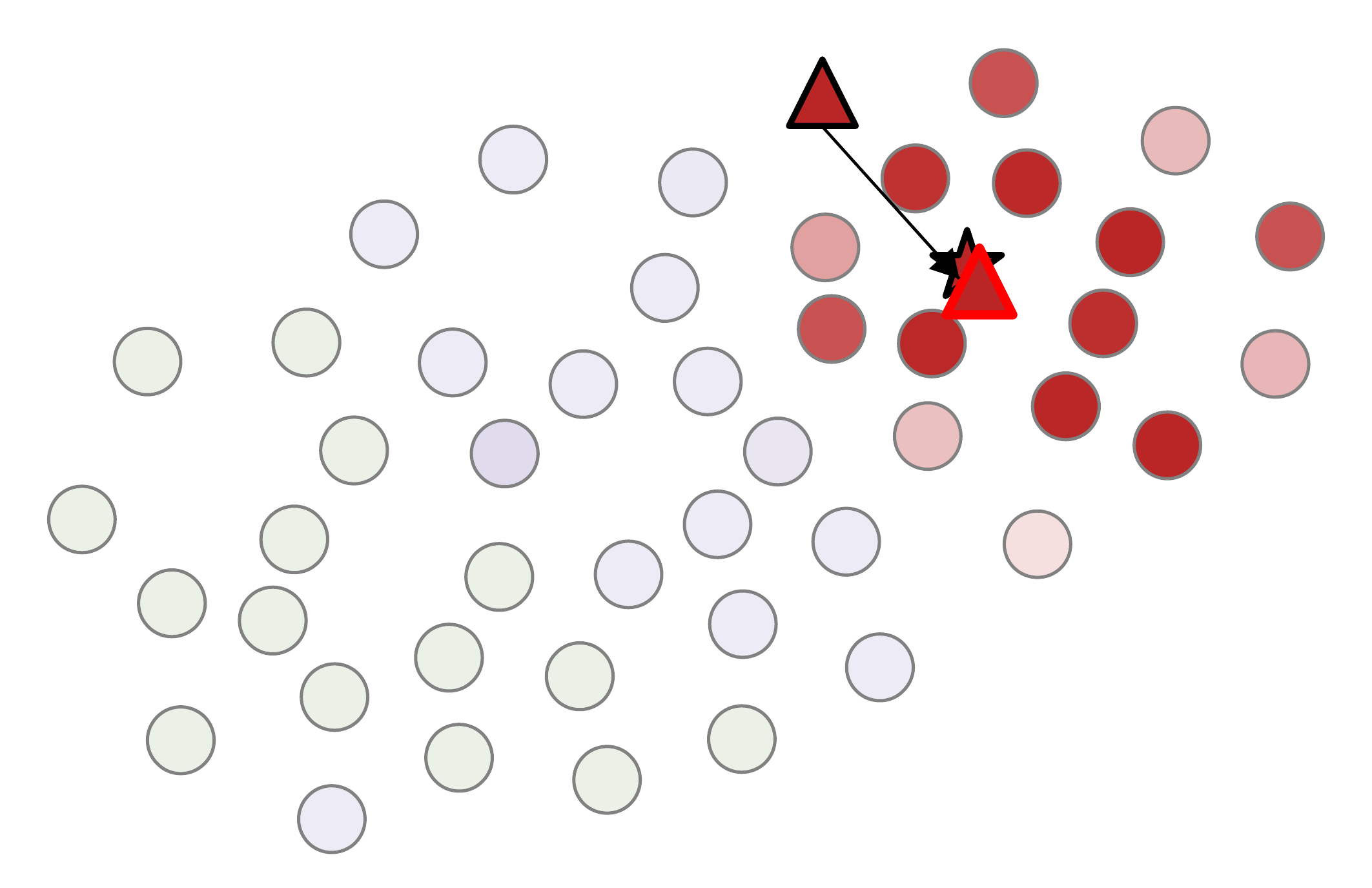}\label{fig:fig1_MCT}}
	\hfill
	\vspace{-0.13in}
    \caption{ \small Transductive inference with confidence scores. We visualize t-SNE embeddings on a 3-way 1-shot task, where each color stands for different class. The numbers show the accuracy increase after transduction for this task. The transparency shows the confidence scores for \emph{red} class.}
	\vspace{-0.20in}
    \label{fig:concept_3way}
\end{figure}

In this work, we aim to tackle this problem by proposing a novel confidence-based transductive inference scheme for metric-based meta-learning models. Specifically, we first propose to meta-learn the distance metric (or metric) to assign different confidence scores to each query (or test) instance for each class, such that the updated prototypes obtained by confidence-weighted averaging of the queries improve classification of the query samples. This is done by learning a metric length-scale term for each individual instance or a pair of instances. However, the confidence prediction on the test instances for \emph{unseen} task should be inevitably unreliable, since the samples come from an unknown distribution. To account for such uncertainties of prediction on an unseen task, we further propose to generate various model and data perturbations, such as random dropping of residual blocks and random augmentations. This randomness helps the model better learn the confidence measure by considering various uncertainties for an unseen task (see Figure~\ref{fig:concept_3way}), and also allows us to take an ensemble over the confidence measures under random perturbations at test time. We refer to this transductive inference using meta-learned input-adaptive confidence under various perturbations as \emph{Meta-Confidence Transduction} (MCT). 

To further enhance the reliability of the output confidence, we introduce additional regularizations to enforce consistency among the transformed samples in the embedding space. Specifically, we compose episodes with differently augmented support and query set and train the model to enforce the distribution of these two sets to be close to each other. Moreover, we also enforce consistency among the dimension-wise classification of the high-dimensional embedding vectors, such that their predictions are coherent. 

We validate our transductive inference scheme for metric-based meta-learning models on four benchmark datasets against existing transductive approaches, which shows that the models using meta-learned confidence significantly outperform existing transductive inference methods, and obtain new state-of-the-art results. We further verify the generality of our MCT on semi-supervised learning tasks, where we assign confidence scores to unlabeled data. The results show that MCT outperforms relevant baselines by large margins, which shows the efficacy of our method. Further ablation studies show that both meta-learning of the input-adaptive distance metric and various perturbations are crucial in the success of our method in assigning correct confidence to each test sample.

Our main contributions are as follows:
\vspace{-0.05in}
\begin{itemize}
\item We propose to \textbf{meta-learn} an \textbf{input-adaptive distance metric}, which allows to output an accurate and reliable confidence for an unseen test samples that can directly improve upon the transductive inference performance.
\vspace{-0.03in}

\item To further enhance the reliability of the learned confidence, we introduce various types of \textbf{model} and \textbf{data perturbations} during meta-learning, such that the meta-learned confidence can better account for uncertainties at unseen tasks.
\vspace{-0.03in}

\item We suggest \textbf{consistency regularizations} across different perturbations and predictions for each embedding dimension, which improves the consistency of the embeddings.
\item We validate our model on four benchmark datasets for few-shot classification and achieve \textbf{new state-of-the-art} results, largely outperforming all baselines. Further experimental validation of our model on semi-supervised few-shot learning also verifies its efficacy.
\end{itemize}

    \vspace{-0.15in}
\section{Related Work}
\vspace{-0.05in}
\paragraph{Distance-based meta-learning for few-shot classification} The goal of few-shot classification is to correctly classify query set examples given only a handful of support set examples. Due to its limited amount of data, each task-specific classifier should resort to the meta-knowledge accumulated from the previous tasks, which is referred to as meta-learning~\cite{thrun98}. Meta-learning of few-shot classification can roughly be divided into several categories such as optimization-based method~\cite{finn2017model,grant2018recasting,lee2018gradient,opt_as_model,leo,cavia}, distance-based approaches~\cite{snell2017prototypical,sung2018learning,vinyals2016matching}, class or task-wise network modulation with amortization~\cite{gordon2018meta,cnaps}, or some combination of those approaches~\cite{das2019two,l2b,oreshkin2018tadam,triantafillou2019meta}. We use a distance-based approach in this work, which allows us to directly compare distance between examples on a metric space. For example, Matching Networks~\cite{vinyals2016matching} use cosine distance, whereas Prototypical Networks~\cite{snell2017prototypical} use euclidean distance with each class prototype set to the mean of support embeddings. 
% More recent approaches propose to set the class prototypes to some global parameters~\cite{gidaris2018dynamic,oreshkin2018tadam,qi2018low}, such that those global prototypes make the learning more stable than the local prototypes differently computed for each episode.
\vspace{-0.15in}
\paragraph{Transductive learning}
Since few-shot classification is intrinsically challenging, we may assume that we can access other unlabeled query examples, which is called transductive learning~\cite{vapnik98}. Here we name a few recent works. TPN~\cite{liu2018learning} constructs a nearest-neighbor graph and propagate labels to pseudo-label the unlabeled query examples. EGNN~\cite{kim2019edge} similarly constructs a nearest-neighbor graph, but utilizes both edge and node features in the update steps. On the other hand, Hou \textit{et al.}~\cite{hou2019cross} tries to update class prototypes by picking top-$k$ confident queries with their own criteria. Our approach also updates class prototypes for each transduction step, but makes use of all the query examples instead of a small subset of $k$ examples.
\vspace{-0.15in}
\paragraph{Semi-supervised learning} In the few-shot classification, semi-supervised learning can access additional large amount of unlabeled data. Ren \textit{et al.}~\cite{ren2018meta} proposed several variants of soft $k$-means method in prototypical networks~\cite{snell2017prototypical}, where soft label is predicted confidence of unlabeled sample. Li \textit{et al.}~\cite{l2ST} proposed the self-training method with pseudo labeling module based on gradient descent approaches~\cite{finn2017model, MTL}. Basically, if an unlabeled query set is used for few-shot classification instead of an additional unlabeled set, it becomes transductive learning, and vice versa. Our approach has connection to soft $k$-means method of Ren \textit{et al.}~\cite{ren2018meta}, but we predict the confidence with input-adaptive distance metric and use meta-learned confidence under various perturbations.
    \vspace{-0.05in}
\section{Preliminaries}
% transductive inference algorithm for metric-based few-shot learning methods, which iteratively updates the class prototypes by taking a weighted sum of the query examples based on their meta-learned confidence score. 
% We also introduce a simple yet effective pixel-wise dense classification loss that is coupled with the conventional instance-wise classification, which we empirically found to be superior over existing models. 

\vspace{-0.05in}
\subsection{Few-shot Classification}
\vspace{-0.05in}
We start by introducing notations. In the conventional $C$-way $N$-shot classification, we first sample $C$ classes randomly from the entire set of classes, and then sample $N$ and $M$ examples from each class for the support set and query set, respectively. We define this sampling distribution as $p(\tau)$. As a result, we have a support set $\mathcal{S} = \{(\bx_i,y_i)\}_{i=1}^{C \times N}$ and query set $\mathcal{Q}=\{(\btx_i,\bty_i)\}_{i=1}^{C \times M}$, where $y,\bty \in \{1,\dots,C\}$ are the class labels. If some portion of the support set is unlabeled, then the problem becomes semi-supervised learning. The convention for the evaluation of few-shot classification models is to use $N\in\{1,5\}$ (i.e. $1$- or $5$-shot) and $M=15$.

The goal of few-shot classification is to correctly classify query examples in $\mathcal{Q}$ given the support set $\mathcal{S}$. Since $\mathcal{S}$ includes only a few examples for each class, conventional learning algorithms will mostly fail due to overfitting (e.g. consider 1-shot classification). Thus, most existing approaches tackle this problem by meta-learning over a task distribution $p(\tau)$, such that the later tasks can benefit from the knowledge obtained over the previous training episodes. 

One of the most popular and successful approaches for few-shot classification is the metric-based approach, in which we aim to learn an embedding function
${f}_{\theta}(\bx) \in \mathbb{R}^{l}$ that maps an input $\bx$ to a latent embedding $\bz$ in an $l$-dimensional metric space (which is usually the penultimate layer of a convolutional network). Support set and query examples are then mapped into this space, such that we can measure the distance between class prototypes and query embeddings.

\vspace{-0.05in}
\subsection{Transductive Inference with Soft $k$-means}
\vspace{-0.05in}
% Yet, even with the meta-learning strategy, few-shot learning remains very challenging, and some existing work has proposed to tackle the problem using \emph{transductive inference}. In transductive inference, when we classify a query example $\btx_1$, we assume that we can access other unlabeled query examples $\btx_2,\dots,\btx_{C\times M}$, and then make use of the intrinsic structure of the data (e.g. top-$k$ nearest neighbors, or nearest neighbor graphs) such that the prediction for each instance gets influenced by the prediction on the other instances that are related to it. 

We now describe and discuss transductive inference using the confidence scores of query examples computed by soft $k$-means algorithm~\cite{ren2018meta}. Suppose that we are given an episode consisting of support set $\mathcal{S}$ and query set $\mathcal{Q}$. We also define $\mathcal{S}_c$ as the set of support examples in class $c$ and $\mathcal{Q}_x = \{\btx_1,\dots,\btx_{C \times M}\}$ as the set of all query instances. Starting from prototypical networks~\cite{snell2017prototypical}, we first compute the initial prototype $P_c^{(0)} = \frac{1}{|\mathcal{S}_c|} \sum_{x \in \mathcal{S}_c} f_\theta(\bx)$ for each class $c = 1,\dots,C$.
%\begin{align}
%\end{align}
Then, for each step $t=1,\dots,T$, and for each query example $\btx \in Q_x$, we compute its confidence score, which denote the probability of it belonging to each class $c$, as follows:
\begin{align}
q_{c}^{(t-1)}(\btx) = 
\frac{\exp(-d(f_\theta(\btx), P_c^{(t-1)} ))}
{\sum_{c'=1}^C \exp(-d(f_\theta(\btx), P_{c'}^{(t-1)}))}
\label{eq:conf}
\end{align}
where $d(\cdot,\cdot)$ is Euclidean distance and $P^{(t-1)}$ denotes $t-1$ steps updated prototype.
We then update the prototypes of class $c$ based on the confidence scores (or soft labels) $q_c^{(t-1)}(\btx)$ \emph{for all} $\tilde{\bx} \in \mathcal{Q}_x$:
\begin{align}
P_c^{(t)}=\frac{\sum_{\bx \in \mathcal{S}_c}1 \cdot f_\theta(\bx)+\sum_{\btx \in \mathcal{Q}_x}q_c^{(t-1)}(\btx) \cdot f_\theta(\btx)}{\sum_{\bx \in \mathcal{S}_c}1+\sum_{\btx \in \mathcal{Q}_x}q_c^{(t-1)}(\btx)}
\label{eq:update}
\end{align}
which is the weighted average that we previously mentioned.
%, analogously to Eq.~\eqref{eq:weighted_average}. 
Note that the confidence of the support examples is always 1, since their class labels are observed. %Note that the framework becomes semi-supervised inference if we use additional set of unlabeled support examples.
We repeat the process until $t=1,\dots,T$.

\vspace{-0.05in}
\paragraph{Questions} However, confidence-based transduction, such as soft $k$-means, leads to a couple of new questions, which is the focus of this work: \emph{1) Is using the confidence of the model indeed helpful in transductive inference?} \emph{2) Can we trust the model confidence that is output from the few-shot task?}

\vspace{-0.05in}
\section{Approach}

\subsection{Meta-Confidence Transduction}
\begin{figure}[t]
\centering
\includegraphics[width=1\linewidth]{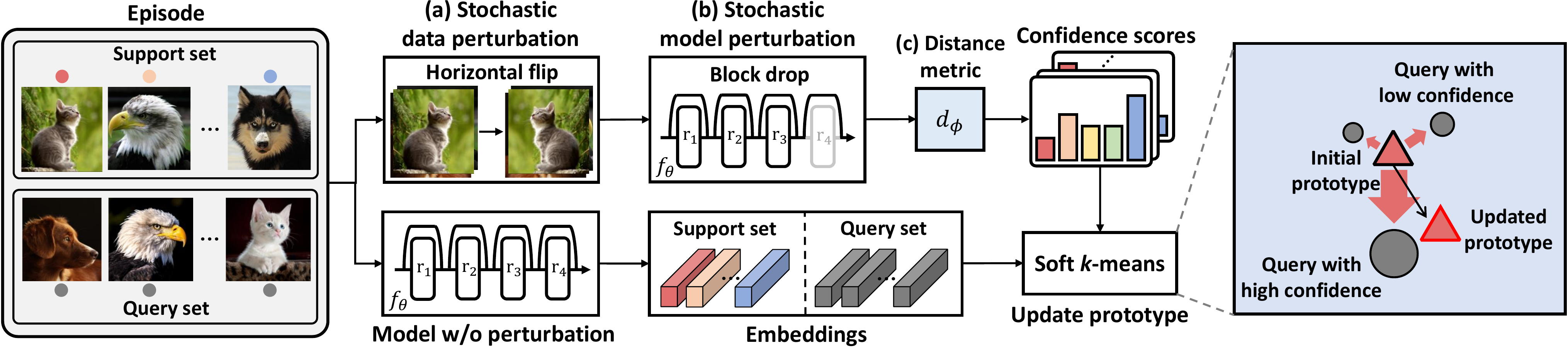}
\vspace{-0.20in}
\caption{\small \textbf{Overview.} (a) To capture data uncertainty, we randomly apply horizontal flip augmentation to the whole data in episode. (b) Along with data uncertainty, we randomly drop the last residual block to capture the model uncertainty. (c) In order to efficiently train the confidence under these perturbations, we \emph{meta-learn} the input-adaptive distance metric.}
\vspace{-0.2in}
\label{fig:overview}
\end{figure}

\vspace{-0.05in}
In order to address the first question, we propose to \emph{Meta-Confidence Transduction} (MCT). As shown in the method overview in Figure~\ref{fig:overview}, we \emph{meta-learn} the distance metric by learning an \emph{input-dependent} temperature scaling for confidence, using the various perturbations on confidence in training.

\vspace{-0.05in}
\paragraph{Meta-learning confidence with input-adaptive distance metric}
We first propose to meta-learn the input-adaptive metric by performing transductive inference during training with query instances, to obtain a metric that yield performance improvements when performing transductive inference using it. Specifically, we meta-learn the distance metric $d_\phi$ in Eq.~\eqref{eq:metric}, which we define as Euclidean distance with normalization and instance-wise metric scaling $g_\phi^I$, or pair-wise metric scaling $g_\phi^P$:
\begin{align}
    d^I_\phi(\mathbf{a}_1, \mathbf{a}_2) = \left\|\frac{\mathbf{a}_1/\|\mathbf{a}_1\|_2}{g^I_\phi(\mathbf{a}_1)} - \frac{\mathbf{a}_2/\|\mathbf{a}_2\|_2}{g^I_\phi(\mathbf{a}_2)}\right\|_2^2,\quad     
    d^P_\phi(\mathbf{a}_1, \mathbf{a}_2) = 
    \left\|\frac{\mathbf{a}_1/\|\mathbf{a}_1\|_2}{g^P_\phi(\mathbf{a}_1,\mathbf{a}_2)} - \frac{\mathbf{a}_2/\|\mathbf{a}_2\|_2}{g^P_\phi(\mathbf{a}_1,\mathbf{a}_2)}\right\|_2^2 
\label{eq:metric}
\end{align}
for all $\mathbf{a}_1,\mathbf{a}_2 \in \mathbb{R}^l$. Note that the normalization allows the confidence to be mainly determined by metric scaling. In order to obtain the optimal scaling function $g_\phi \in \{g_\phi^I,g_\phi^P\}$ for transduction, we first compute the query likelihoods after $T$ transduction steps, and then optimize $\phi$, the parameter of the scaling function $g_\phi$ by minimizing the following instance-wise loss for $d_\phi \in \{d_\phi^I, d_\phi^P\}$:
\begin{align}
L_I^\tau(\theta,\phi) &= \frac{1}{|\mathcal{Q}|}\sum_{(\btx,\bty) \in \mathcal{Q}} -\log p(\bty|\btx, \mathcal{S};\theta,\phi) \\
&= \frac{1}{|\mathcal{Q}|}\sum_{(\btx,\bty) \in \mathcal{Q}} \left\{d_\phi(f_\theta(\btx), P_c^{(T)} )
+ \sum_{c'=1}^C \exp(-d_\phi(f_\theta(\btx), P_{c'}^{(T)}))\right\}.
\label{eq:inst_loss}
\end{align}
As for $g_{\phi}$, we simply use a CNN with fully-connected layers which takes either the feature map of an instance or the concatenated feature map of a pair of instances as an input. We set the number of transduction steps to $T=1$ for training to minimize the computational cost, but use $T=10$ for test.

\begin{algorithm}[t!]
  \caption{Meta-learning confidence with model and data perturbation.}
  \small
    \begin{flushleft}
        \textbf{Require:} The set of support examples $\mathcal{S}_c$, for each class $c \in \{1,\dots,C\}$.
        \\
        \textbf{Require:} The set of all query examples $(\btx,\bty) \in \mathcal{Q}$.
        \\
        \textbf{Require:} Full-path embedding function $f_\theta$ and block-dropped embedding function $f^{D}_\theta$.
        \\
        \textbf{Require:} Flip augmentation Aug($\cdot$) and define $f^{A}_{\theta}$ as $f_{\theta}(\text{Aug}(\cdot))$.
        \\
    \end{flushleft}
  \begin{algorithmic}[1]
    
    \State $h_{\theta} \gets \text{Sample from }  \{f_{\theta}, 
    f^{D}_{\theta}, 
    f^{A}_{\theta}, 
    f^{A,D}_{\theta}
    \} $ \Comment{Select a confidence space}
    
    \For{$c \in \{1,\dots,C\}$}
        \State $P'_{c} \gets \frac{1}{|\mathcal{S}_c|} \sum_{\bx \in \mathcal{S}_c} h_\theta(\bx)$. \Comment{Compute prototype on confidence space}
    \EndFor
    
    \For{$c \in \{1,\dots,C\}$}
        \State $q_c(\btx) \gets \cfrac{\exp \bigl( -d(h_\theta(\btx), P'_{c})\bigr)}
        {\sum_{c'=1}^C \exp\bigl(-d(h_\theta(\btx), P'_{c'})\bigr)}$ for all $\btx \in \mathcal{Q}_x$
        \Comment{Compute confidence score}
        \State $P_{c} \gets \cfrac{\sum_{\bx \in \mathcal{S}_c}1 \cdot f_{\theta}(\bx)+\sum_{\btx \in \mathcal{Q}_x}q_c(\btx) \cdot f_{\theta}(\btx)}{\sum_{\bx \in \mathcal{S}_c}1+\sum_{\btx \in \mathcal{Q}_x}q_c(\btx)}$ \Comment{Compute prototype on full-path space}
    \EndFor
    \State $J \gets 0$ \Comment{Initialize loss}
    % \For{$c \in \{1,\dots,C\}$}
    \For{$(\btx,\bty) \in \mathcal{Q}$}
        \State
        $J \gets J+\cfrac{1}{|\mathcal{Q}_x|}\Bigg[d(f_{\theta}(\btx),P_{\bty})+\text{log}\displaystyle \sum_{c'}\text{exp}\bigl(-d(f_{\theta}(\btx),P_{c'})\bigr)\Bigg]$  \Comment{Update loss}
        % \EndFor
    \EndFor
  \end{algorithmic}
  \label{algo:train_MCT}
\end{algorithm}
\vspace{-0.1in}
\paragraph{Model and data perturbations}
The model confidence from few-shot tasks is intrinsically unreliable due to the data scarcity problems, even if the model has been meta-learned over similar tasks.
One way to output more reliable and consistent confidence scores is to enforce the model to output consistent predictions while perturbing either the model or the data.
In this work, we consider the following two sources of perturbations:
\vspace{-0.05in}
\begin{itemize}
    \item \textbf{Model perturbation:} We consider two confidence scores, one from the full network (full-path) and the other from a sub-network generated by dropping a block (drop-path) ~\cite{dropout_as_bayesian,veit2016residual,blockdrop} from the full network.
    \item \textbf{Data perturbation:} We also consider two confidence scores, one from the original image and the other from horizontally flipped image.
\end{itemize}
\vspace{-0.05in}
By jointly considering these two sources of perturbations, we can have a total of four ($2 \times 2$) scenarios (or sources) of possible transductive inferences. As shown in Algorithm \ref{algo:train_MCT}, at training time, we randomly select a source of confidence and simulate a single transduction step. However at test time, we perform transductive inference for all scenarios using the ensemble confidence obtained from all perturbed sources. This process is done $T$ times to get the final confidence scores. By doing so, we can enforce the model to consistently perform well under various transduction scenarios with different perturbations, leading to better performance due to the ensemble effect of meta-learned confidences (see the Section~\ref{app:C} of the appendix for more details).

\vspace{-0.05in}

\subsection{Consistency Regularization}\label{sec:consist}
\vspace{-0.05in}
In order to address the second question, we suggest consistency regularization for data and embedding. The quality of the confidence scores can be improved with consistency regularization. In semi-supervised learning, consistency regularization is one of the most popular techniques as it allows for unlabeled examples to output consistent predictions under various perturbations, thereby improving the quality of the confidence scores~\cite{temporal, fixmatch,mean_teacher}.

\paragraph{Consistency over data perturbation}
We also propose to enforce the model to output consistent predictions under various perturbations of the query examples. The idea is that, even though we perturb the query examples by large amount, a good and discriminative embedding space for transductive inference should be able to correctly classify them.
Specifically, we apply only the horizontal flipping and shifting to the examples in the support set (\emph{weak} augmentation), whereas we apply horizontal flipping, shifting, RandAugment~\cite{randaugment} and CutOut~\cite{cutout} to the examples in the query set (\emph{strong} augmentation), and perform classification with those augmentations. This approach is related to FixMatch~\cite{fixmatch} algorithm for semi-supervised learning, but we apply various augmentations to disjoint sets rather than to the same instance, which allows to achieve the same effect as a regularization without an explicit consistency loss.

% In general, human can classify the data of the same class into the same class even if the images are different or transformed. In other words, human can have almost same confidence in data of same class regardless of data. In order to make the model has such capability, many previous works proposed various regularization methods such as data augmentation and consistency loss. 
%We also enforce the \emph{any} images for each class can be projected in 
%adjacent embedding space. 
%Our strategy combines data augmentation with prototypical networks. 
%In prototypical networks, we make the class representation by using support set. 
%hen, we train query set to be closer to support set. 
%By doing so, we can make the model have close distribution between them in metric space. 
%To regularize model, we enlarge distribution of these two sets. Toward this goal, 
%we apply different image augmentation to support and query set.
%In general case, we apply same data augmentation to both of them. However, in our approach, we further transform query set to move away from data distribution of support set.
% Considering opposite case, we found that large variance of prototype for each task makes the training unstable.
%We refer to set of horizontal flip and shift as `weak', and `strong' indicates that RandAugment~\cite{randaugment} and Cutout~\cite{cutout} are additionally applied to `weak'. 

% In addition, training would be much faster, since each input passes through the network only once.

\vspace{-0.05in}
\paragraph{Consistency over dimensions of embedding space}
Dense classification (DC) \cite{lifchitz2019dense} achieves successful performance improvement in few-shot classification. However, they apply spatial pooling to feature maps, in order to make embeddings at testing. This causes unnecessary bottlenecks, making it difficult to completely use the learned spatial information. To alleviate this problem, we reinterpret DC as a regularizer on the high dimensional embedding being learned. In other words, we do not apply spatial pooling at both training and testing, and then use flattened feature map as the embedding for each instance. We found that computing the distance with densely matching the spatial embeddings improves performance, without any additional parameters. When training with DC, we additionally compute dimension-wise loss $L_D^{\tau}$, the average classification loss for each dimension of embedding (e.g. 64-way classification for miniImageNet). Hence, final learning objective is $L=E_{p(\tau)}[\lambda L_I^{\tau}+L_D^{\tau}]$, where $L_I^{\tau}$ is the instance-wise loss in Eq.~\ref{eq:inst_loss} and $\lambda$ is the balancing factor.
    \section{Experiments}
\paragraph{Dataset} We validate our method on four popular benchmark datasets for few-shot classification. \textbf{1) miniImageNet}, \textbf{2) tieredImageNet}, \textbf{3) CIFAR-FS}, and \textbf{4) FC100}. Please see the Section~\ref{app:A.1} of the appendix regarding the detailed information for each of the datasets.

% \textbf{1) miniImageNet.} This dataset~\cite{vinyals2016matching} consists of a subset of $100$ classes sampled from the ImageNet dataset~\cite{russakovsky2015imagenet}. Each class has $600$ images, resized to $84\times84$ pixels. We use the split of $64/16/20$ for training/validation/test.

% \textbf{2) tieredImageNet.} This dataset~\cite{ren2018meta} is another subset of ImageNet, that consists of $779,165$ images of $84\times{84}$ pixels collected from $608$ classes. The task is to generalize the few-shot classifier over $34$ different superclasses. Thus the entire dataset is split into $20/6/8$ superclasses for training/validation/test, where each superclass contains $351$, $97$, and $160$ low-level classes respectively.  

% \textbf{3) CIFAR-FS.} This dataset~\cite{bertinetto2018metalearning} is a variant of CIFAR-100 dataset used for few-shot classification, which contains $100$ classes that describe general object categories. For each class, there are $600$ images of $32\times{32}$ pixels. The dataset is split into $64/16/20$ classes for training/validation/test.

% \textbf{4) FC100.} This is another few-shot classification dataset~\cite{oreshkin2018tadam} compiled by reorganizing the CIFAR-100 dataset. The task for this dataset is to generalize across $20$ superclasses, as done with the tieredImageNet dataset. The superclasses are divided into $12/4/4$ classes for training/validation/test, each of which contains $60/20/20$ low-level classes, respectively.
% \vspace{-0.05in}
\begin{table*}[t!]
\centering
\footnotesize
\vspace{-0.05in}
\caption{\small\textbf{Average classification performance} over 1000 randomly generated episodes, with 95\% confidence intervals. We consider 5-way classification on all the datasets. $^*$ denotes it is reported from \cite{DPGN}.}
\vspace*{0.03in}
\small
 \resizebox{1.\linewidth}{!}{
 \begin{tabular}{ c c c c c c c }
 \hline
& \multirow{2}{*}{Model}  &
\multirow{2}{*}{Backbone} &
\multicolumn{2}{c}{miniImageNet}& \multicolumn{2}{c}{tieredImageNet} \\
 &  &  & 1-shot & 5-shot & 1-shot & 5-shot \\
 \hline
 \hline
 \multirow{8}{*}{Inductive}&
MTL$^{*}$~\cite{MTL} & ResNet-12 &
61.20\tiny$\pm$1.80 & 75.53\tiny$\pm$0.80 &
65.62\tiny$\pm$1.80 & 80.61\tiny$\pm$0.90 \\

& TapNet~\cite{yoon2019tapnet} & ResNet-12    & 61.65\tiny$\pm$0.15 & 76.36\tiny$\pm$0.10 & 63.08\tiny$\pm$0.15 & 80.26\tiny$\pm$0.12 \\

&TADAM~\cite{oreshkin2018tadam} & ResNet-12    &
58.56\tiny$\pm$0.39 & 76.65\tiny$\pm$0.38 &
62.13\tiny$\pm$0.31 & 81.92\tiny$\pm$0.30  \\

& MetaOpt-SVM~\cite{lee2019meta} & ResNet-12    & 62.64\tiny$\pm$0.61 & 78.63\tiny$\pm$0.46 &
65.99\tiny$\pm$0.72 & 81.56\tiny$\pm$0.53 \\

& Dense~\cite{lifchitz2019dense}  & ResNet-12    & 61.26\tiny$\pm$0.20 & 79.01\tiny$\pm$0.13 &
- & - \\

& CAN~\cite{hou2019cross} & ResNet-12    &
63.85\tiny$\pm$0.48 & 79.44\tiny$\pm$0.34 & 
\textbf{69.89\tiny$\pm$0.51} & 84.23\tiny$\pm$0.37 \\

\rowcolor{black!10} \cellcolor{black!0} & MCI (Pair) & ResNet-12 & 
64.49\tiny$\pm$0.64 &
81.63\tiny$\pm$0.44 &
68.41\tiny$\pm$0.73 &
84.60\tiny$\pm$0.50 \\

\rowcolor{black!10} \cellcolor{black!0} &  MCI (Instance) & ResNet-12 & 
\textbf{65.34\tiny$\pm$0.63} &
\textbf{82.15\tiny$\pm$0.45} &
69.66\tiny$\pm$0.72 &
\textbf{85.29\tiny$\pm$0.49} \\

\hline

\multirow{10}{*}{Transductive} & TPN~\cite{liu2018learning} & ConvNet-64   & 55.51\tiny$\pm$0.86 & 69.86\tiny$\pm$0.65 &
59.91\tiny$\pm$0.94 & 73.30\tiny$\pm$0.75 \\

& EGNN$^{*}$~\cite{kim2019edge} & ConvNet-256  &
59.63\tiny$\pm$0.52  & 76.34\tiny$\pm$0.48  &
63.52\tiny$\pm$0.52  & 80.24\tiny$\pm$0.49 \\

& TEAM~\cite{trans_episodic} & ResNet-18    &
60.07 & 75.90 &
- & - \\

& MAML+SCA~\cite{SCA} & DenseNet  &
62.86\tiny$\pm$0.79 & 77.46\tiny$\pm$1.18 &
- & - \\

& Fine-tuning~\cite{baseline} & WRN-28-10  &
65.73\tiny$\pm$0.68 & 78.40\tiny$\pm$0.52 &
73.34\tiny$\pm$0.71 & 85.50\tiny$\pm$0.50 \\

& SIB~\cite{SIB} & WRN-28-10  &
70.0\tiny$\pm$0.6 & 79.2\tiny$\pm$0.4 &
- & - \\

& CAN + Top-$k$~\cite{hou2019cross} & ResNet-12    & 67.19\tiny$\pm$0.55 & 80.64\tiny$\pm$0.35 &
73.21\tiny$\pm$0.58 & 84.93\tiny$\pm$0.38 \\

& DPGN~\cite{DPGN} & ResNet-12    &
67.77\tiny$\pm$0.32 &
84.60\tiny$\pm$0.43 &
72.45\tiny$\pm$0.51 &
87.24\tiny$\pm$0.39 \\

\rowcolor{black!10} \cellcolor{black!0} & MCT (Pair) & ResNet-12    &
76.16\tiny$\pm$0.89 &
85.22\tiny$\pm$0.42 &
80.68\tiny$\pm$0.89 &
86.63\tiny$\pm$0.89 \\

\rowcolor{black!10} \cellcolor{black!0} & MCT (Instance) & ResNet-12    &
\textbf{78.55\tiny$\pm$0.86} &
\textbf{86.03\tiny$\pm$0.42} &
\textbf{82.32\tiny$\pm$0.81} &
\textbf{87.36\tiny$\pm$0.50} \\
\hline
 \end{tabular}
 }
 \vspace{-0.1in}
 \label{tbl:mini_tiered}
\end{table*}
% \begin{tabular}{c c c c c c c c c c c c}
\begin{table*}[t!]
\centering
\footnotesize
\vspace{-0.15in}
\caption{\small\textbf{Average classification performance} on CIFAR-FS and FC100.}
\vspace*{0.03in}
\small
 \resizebox{1.\linewidth}{!}{
 \begin{tabular}{ c c c c c c c }
 \hline
& \multirow{2}{*}{Model}  &
\multirow{2}{*}{Backbone} &
\multicolumn{2}{c}{CIFAR-FS} &
\multicolumn{2}{c}{FC100} \\

 &  &  & 1-shot & 5-shot & 1-shot & 5-shot \\
 \hline
 \hline
\multirow{5}{*}{Inductive}&
TADAM~\cite{oreshkin2018tadam} & ResNet-12    &
- & - &
40.1\tiny$\pm$0.4 & 56.1\tiny$\pm$0.4 \\

& MetaOpt-SVM~\cite{lee2019meta} & ResNet-12    & 
72.00\tiny$\pm$0.70 & 84.20\tiny$\pm$0.50 &
41.10\tiny$\pm$0.60 &  55.50\tiny$\pm$0.60\\

& Dense~\cite{lifchitz2019dense}  & ResNet-12    &
 - & - & 
42.04\tiny$\pm$0.17 & 57.05\tiny$\pm$0.16\\

\rowcolor{black!10} \cellcolor{black!0} & MCI (Pair) & ResNet-12 & 
76.23\tiny$\pm$0.72 &
88.39\tiny$\pm$0.44 &
43.01\tiny$\pm$0.58 &
59.67\tiny$\pm$0.56 \\

\rowcolor{black!10} \cellcolor{black!0} & MCI (Instance) & ResNet-12 & 
\textbf{77.84\tiny$\pm$0.64} &
\textbf{89.11\tiny$\pm$0.45} &
\textbf{44.69\tiny$\pm$0.60} &
\textbf{60.33\tiny$\pm$0.59} \\
\hline

\multirow{5}{*}{Transductive}
& Fine-tuning~\cite{baseline} & WRN-28-10  &
76.58\tiny$\pm$0.68 & 85.79\tiny$\pm$0.50 &
43.16\tiny$\pm$0.59 & 57.57\tiny$\pm$0.55 \\

& SIB~\cite{SIB} & WRN-28-10  &
80.0\tiny$\pm$0.6 & 85.3\tiny$\pm$0.4 &
-&-\\

& DPGN~\cite{DPGN} & ResNet-12    &
77.90\tiny$\pm$0.50 &
90.20\tiny$\pm$0.40 &
-&-\\

\rowcolor{black!10} \cellcolor{black!0} & MCT (Pair) & ResNet-12    &
\textbf{87.28\tiny$\pm$0.70} &
\textbf{90.50\tiny$\pm$0.43} &
\textbf{51.27\tiny$\pm$0.80} &
62.59\tiny$\pm$0.60 \\

\rowcolor{black!10} \cellcolor{black!0} & MCT (Instance) & ResNet-12    &
85.61\tiny$\pm$0.69 &
90.03\tiny$\pm$0.46 &
51.16\tiny$\pm$0.88 &
\textbf{63.28\tiny$\pm$0.61} \\
\hline
 \end{tabular}
 }
 \vspace{-0.15in}
 \label{tbl:cifar_fc100}
\end{table*}

\vspace{-0.05in}
\paragraph{Experimental setting} Here we mention a few important experimental settings of our model. During training, we apply the weight decay of $0.0005$, and unless otherwise indicated, apply the augmentations proposed in Section~\ref{sec:consist} by default. When the image size is $32\times 32$, we apply max-pooling only to the second and the fourth layer to increase the dimensionality of the final embedding space. For our full models, we evaluate the expectation over task distribution $p(\tau)$ via Monte-Carlo (MC) approximation with a single sample during training to obtain the learning objective, where we set $\lambda=0.5$ which we found with a validation set. More details (e.g. learning rate scheduling, detailed network architectures and settings for semi-supervised experiment) can be found in the Section~\ref{app:A} and~\ref{app:B} of the appendix.

\subsection{Main Results}
\paragraph{Inductive inference} 
We first examine the results of inductive inference. We define \emph{Meta-Confidence Induction (MCI)} as an our proposed metric with consistency regularizations only. The top rows of Table~\ref{tbl:mini_tiered} and Table~\ref{tbl:cifar_fc100} show the accuracy of MCI and the existing inductive inference methods for few-shot classification. Our model achieves \textbf{new state-of-the-art results} for inductive inferecne models on all four benchmark datasets with significant margins. This performance gain is coming from both the consistency regularization over the data perturbation and on the dimensions of the embeddings. We analyze each component in detail in the following sections.

\begin{wraptable}{t}{0.50\textwidth}
\begin{center}
	\vspace{-0.08in}
	\resizebox{1\linewidth}{!}{
		\small
		\begin{tabular}{cccc}
			\toprule
			\multirow{2}{*}{Model} &
            \multirow{2}{*}{Backbone} &
            \multicolumn{2}{c}{miniImageNet} \\
             &  & 1-shot & 5-shot \\
             \hline
             \hline
            TPN~\cite{liu2018learning} & 
            ConvNet-64 &
            55.51\tiny$\pm$0.86 &
            69.86\tiny$\pm$0.65\\
            MCT (Instance) & 
            ConvNet-64 &
            \textbf{63.53\tiny$\pm$0.91} &
            \textbf{75.15\tiny$\pm$0.56}\\
            \hline
            EGNN~\cite{kim2019edge} & 
            ConvNet-256 &
            59.63\tiny$\pm$0.52 &
            76.34\tiny$\pm$0.48 \\
            MCT (Instance) & 
            ConvNet-256 &
            \textbf{70.10\tiny$\pm$0.87} &
            \textbf{80.56\tiny$\pm$0.49}\\
            % MetaConf-T (Instance-wise) & 
            % ConvNet-256 &
            % 69.32\tiny$\pm$0.89 &
            % 79.89\tiny$\pm$0.49 \\
            %  \hline
            % TPN~\cite{liu2018learning} &
            % ResNet-12 &
            % 59.46 & 75.65 \\
            % DPGN~\cite{DPGN} & 
            % ResNet-12 &
            % 67.77\tiny$\pm$0.32 &
            % 84.60\tiny$\pm$0.43 \\
            % % MCT (Instance)$^{\dagger}$ & 
            % % ResNet-12 &
            % % 76.65\tiny$\pm$0.84 &
            % % 85.09\tiny$\pm$0.43 \\
            % MCT (Instance) & 
            % ResNet-12 &
            % \textbf{78.55\tiny$\pm$0.86} &
            % \textbf{86.03\tiny$\pm$0.42} \\
			\bottomrule	
		\end{tabular}
	}
	\vspace{-0.10in}
	\caption{\small Comparison with other transductive models.}\label{tbl:ablation_transduction}
	\vspace{-0.15in}
\end{center}
\end{wraptable}

%  $^{\dagger}$: model without our proposed augmentation.
\paragraph{Transductive inference} 
The bottom rows of Table~\ref{tbl:mini_tiered} and Table~\ref{tbl:cifar_fc100} show the results of transductive inference with the baselines and our full model, Meta-Confidence Transduction (MCT), which performs transductive inference with the meta-learned confidence. We again achieve \textbf{new state-of-the-art results} on all the datasets, with particularly good performance on one-shot classification. 
%To compare other baselines, we divide them into two broad categrories such as graph-based model and non graph-based model. 
%We compare our model against the graph based models such as TPN~\cite{liu2018learning}, EGNN~\cite{kim2019edge} and DPGN~\cite{DPGN}. Our model outperforms all the baselines, including the most recent model DPGN with significant margins for both 1-shot and 5-shot classifications.%Whereas the most recent model,  outperforms TPN and EGNN with significant margin in 5-shot classification, significant performance improvement in 1-shot scenarios. Compared to them, our model significantly outperforms these graph based models in 1-shot scenarios, also including 5-shot. 
%We next compare our model against non graph based models. For the 5-shot setting, our model outperforms other baselines with large margin, despite using a much shallower backbone. In 1-shot scenarios, the gap of the performance is even greater. 
For fair comparsion against TPN~\cite{liu2018learning} and EGNN~\cite{kim2019edge} that use different backbone networks, we further perform an additional experiments using shallow backbone networks in Table~\ref{tbl:ablation_transduction}. Again, our model largely outperforms all baselines. Note that we use MCT without model perturbation (block drop) since ConvNet-64 and ConvNet-256 do not have skip connections. %Furthermore, we can achieve better performance against TPN with ResNet-12.

\begin{table*}[t!]
\centering
\footnotesize
% \vspace{-0.05in}
\caption{\small \textbf{Semi-supervised few-shot classification performance}. We consider 5-way classification on miniImageNet (`mini') and tieredImageNet (`tiered'). The baseline results are drawn from \cite{l2ST}. All results are based on pre-trained ResNet-12 with full dataset in conventional supervised manner. “w/$\mathcal{D}$” means that unlabeled set includes 3 distracting classes, which does not overlap the label space of the support set~\cite{l2ST,liu2018learning,ren2018meta}.}
\vspace*{0.03in}
 \begin{tabularx}{1.0\textwidth}{
 c @{\extracolsep{\fill}}
 c @{\extracolsep{\fill}}
 c @{\extracolsep{\fill}}
 c @{\extracolsep{\fill}}
 c @{\extracolsep{\fill}}
 c @{\extracolsep{\fill}}
 c @{\extracolsep{\fill}}
 c @{\extracolsep{\fill}}
 c @{\extracolsep{\fill}}}
 
 \hline
\multirow{2}{*}{Model}  & \multicolumn{2}{c}{mini} & \multicolumn{2}{c}{tiered}&
 \multicolumn{2}{c}{mini w/$\mathcal{D}$} &
 \multicolumn{2}{c}{tiered w/$\mathcal{D}$} \\
  & 1-shot & 5-shot & 1-shot & 5-shot &
 1-shot & 5-shot & 1-shot & 5-shot\\
 \hline
 \hline
Masked Soft $k$-Means~\cite{ren2018meta} &
62.1 & 73.6 &
68.6 & 81.0 &
61.0 & 72.0 &
66.9 & 80.2 \\

TPN~\cite{liu2018learning} &
62.7 & 74.2 &
72.1 & 83.3 &
61.3 & 72.4 &
71.5 & 82.7 \\

LST~\cite{li2019learning} &
70.1 & 78.7 &
\textbf{77.7} & 85.2 &
64.1 & 77.4 &
73.5 & 83.4 \\

MCT (Instance) &
\textbf{73.8\tiny$\pm$0.7}  &
\textbf{84.4\tiny$\pm$0.5}  &
76.9\tiny$\pm$0.7 & 
\textbf{86.3$\tiny\pm$0.5}  &
\textbf{69.6\tiny$\pm$0.7}  & 
\textbf{81.3\tiny$\pm$0.5}  &
\textbf{74.5\tiny$\pm$0.7}  & 
\textbf{84.0\tiny$\pm$0.5}  \\
\hline
 \end{tabularx}
 \vspace{-0.15in}
 \label{tbl:semi_result}
\end{table*}
\paragraph{Semi-supervised inference} We also perform experiments on semi-supervised classification in Table~\ref{tbl:semi_result} to further validate the effectiveness and generality of our MCT. We follow the same experimental setting described in Li \textit{et al.}~\cite{l2ST}. In the semi-supervised setting, instead of computing the confidence scores of query examples, we compute the confidence scores of unlabeled support examples in order to update the class prototype. Again, our MCT largely outperforms all the baselines including the recent LST model. The results demonstrate the effectiveness of our consistency regularizations and the distance metric scaling for correctly assigning confidence scores to unlabeled examples.

%is because the average confidence computed over various perturbations makes the label of unreliable samples smoother and allows highly confident samples to have sharp labels. This allows us to utilize every samples without selection and get the superior performance.

%LST~\cite{l2ST} model outperforms masked soft $k$-means~\cite{ren2018meta} and TPN~\cite{liu2018learning} with large margin. LST model trains an additional pseudo labeling module that selects the top-$k$ confident samples and assigns soft label of the selected samples. On the other hand our MetaConf-T model significantly outperforms LST even without such an additional module. In our model, average confidence against various perturbation scenarios makes the label of unreliable samples smoother and allows highly confident samples to have sharp labels. This allows us to utilize every samples without selection and get the superior performance.

\subsection{Ablation Studies}
We next perform ablation studies of our model on miniImageNet dataset to identify from where the performance improvements come from. We use prototypical networks (PN) with ResNet-12 backbone networks for these experiments, including only a single component of MCT at a time.
\begin{table*}[t]
\centering
\footnotesize
 \caption{\small\textbf{Average classification performance} over 1000 randomly generated episodes, with 95\% confidence intervals. $d(\cdot,\cdot)$ denotes Euclidean distance. $s \in \mathbb{R}$ is a learnable parameter initialized to $7.5$, following \cite{oreshkin2018tadam}.}
\small
 \resizebox{1.\linewidth}{!}{
 \begin{tabular}{ c c c c c c}
 
 \hline
\multirow{2}{*}{Model} &
\multirow{2}{*}{Distance Metric} &
\multicolumn{2}{c}{Inductive} &
\multicolumn{2}{c}{Transductive} \\
 &  & 1-shot & 5-shot & 1-shot & 5-shot \\
 \hline
 \hline
Prototypical Networks (PN)~\cite{snell2017prototypical} & 
$d(\mathbf{a}_1, \mathbf{a}_2)$ &
57.36\tiny$\pm$0.66 & 75.59\tiny$\pm$0.51 &
68.58\tiny$\pm$0.92 & 78.71\tiny$\pm$0.53 \\

PN + metric scaling~\cite{oreshkin2018tadam} & $s \cdot d(\mathbf{a}_1, \mathbf{a}_2)$ &
55.43\tiny$\pm$0.67 & 74.52\tiny$\pm$0.49 &
68.34\tiny$\pm$0.87 & 78.57\tiny$\pm$0.51 \\

% PN w/ cosine distance + metric scaling & 
% $s \cdot cos(\mathbf{a}_1, \mathbf{a}_2)$    &
% 58.41\tiny$\pm$0.64 & 73.05\tiny$\pm$0.53 &
% 65.82\tiny$\pm$0.89 & 75.99\tiny$\pm$0.55 \\

PN + Instance-wise metric (Eq. \ref{eq:metric}) & 
$d_{\phi}^I(\mathbf{a}_1,\mathbf{a}_2)$&
61.08\tiny$\pm$0.66 &
77.26\tiny$\pm$0.46 &
70.34\tiny$\pm$0.87 &
79.54\tiny$\pm$0.54 \\

PN + Pair-wise metric (Eq. \ref{eq:metric}) & 
$d_{\phi}^P(\mathbf{a}_1,\mathbf{a}_2)$&
\textbf{61.81\tiny$\pm$0.58} &
\textbf{77.67\tiny$\pm$0.50} &
\textbf{71.95\tiny$\pm$0.81} &
\textbf{81.06\tiny$\pm$0.51} \\
\hline
 \end{tabular}
 }
 \vspace{-0.1in}
 \label{tbl:ablation_metric}
\end{table*}

\begin{wrapfigure}{t}{0.36\textwidth}
    \vspace{-0.25in}
    \centering
    \subfigure{\includegraphics[width=0.49\linewidth]{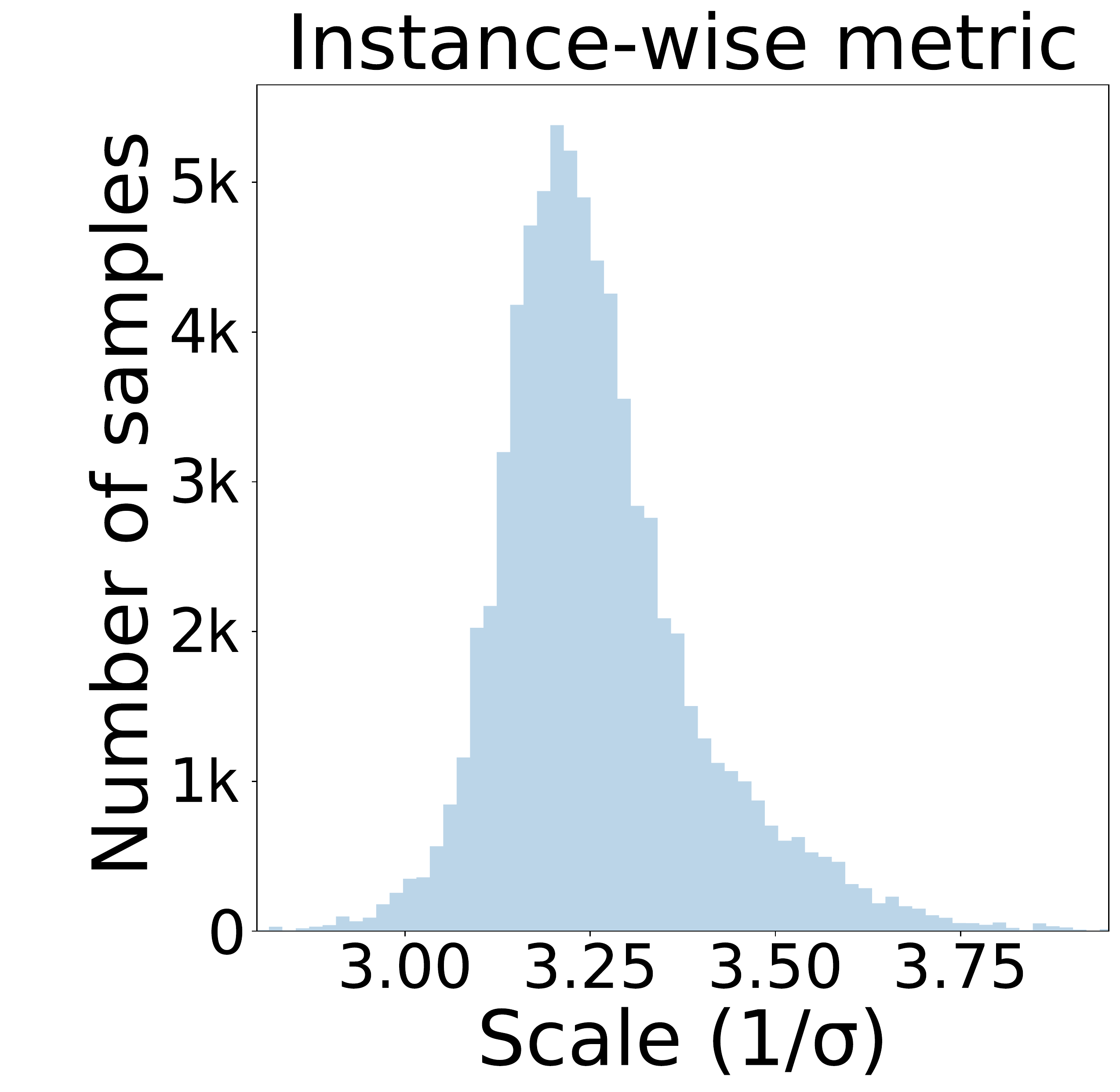}\label{fig:scale_IMS}}
    \subfigure{\includegraphics[width=0.49\linewidth]{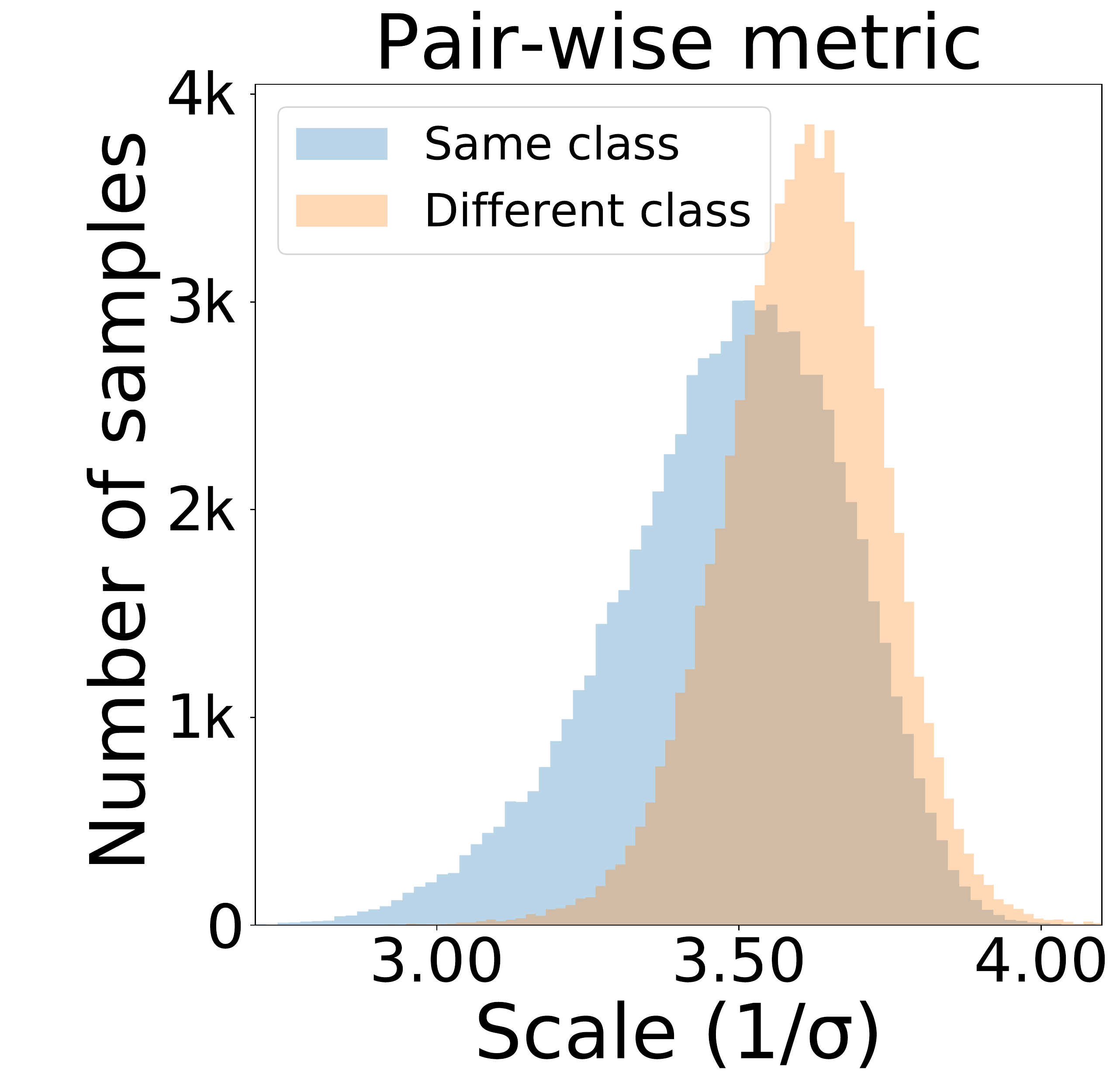}\label{fig:scale_PMS}}
    \vspace{-0.25in}
    \caption{\small Histogram of metric scale, on a miniImageNet 5-way 5-shot task. $\sigma$ corresponds to $g_{\phi}$.}\label{fig:scale_histogram}
    \vspace{-0.10in}
\end{wrapfigure}
\paragraph{Effect of the distance metrics} 
We first study the effect of the distance metric in Table~\ref{tbl:ablation_metric}. The performance in the transductive inference columns correspond to each of the models with the transductive inference with naive soft $k$-means algorithm~\cite{ren2018meta} without model and data perturbations. We see that the PN with metric scaling underperforms the plain PN with Euclidean distance. On the other hand, the proposed instance-wise and pair-wise metric significatly outperform both distances in both inductive and transductive inference settings, demonstrating the effectiveness of our input-dependent metric scaling methods over globally shared metric scaling. In Figure~\ref{fig:scale_histogram}, we observe that instance-wise metric scaling assigns various scales to different inputs, whereas the pair-wise metric scaling assigns low values between the samples from the same class and high values between samples from different classes.

\begin{wraptable}{t}{0.60\textwidth}
\centering
\vspace{-0.12in}

\small

\resizebox{0.6\textwidth}{!}{
 \begin{tabular}{ c c | c  c | c c }

 \hline
Data & Model &
\multicolumn{2}{c|}{miniImageNet 1-shot} &
\multicolumn{2}{c}{miniImageNet 5-shot}  \\
Perturb & Perturb & NLL& Transduction & NLL & Transduction \\
 \hline
 \hline
$\xmark$ & $\xmark$ &
1.11 &
71.95\tiny$\pm$0.81 & 
0.82 &
81.06\tiny$\pm$0.51 \\

$\cmark$ & $\xmark$ &
1.09 &
73.93\tiny$\pm$0.85 &
0.68 &
81.93\tiny$\pm$0.49 \\

$\xmark$ & $\cmark$ &
\textbf{1.04} &
74.07\tiny$\pm$0.85 &
\textbf{0.60} &
 82.62\tiny$\pm$0.47 \\

$\cmark$ & $\cmark$ &
1.09 &
\textbf{74.73\tiny$\pm$0.86} &
\textbf{0.60} &
\textbf{83.36\tiny$\pm$0.45} \\
\hline
 \end{tabular}
 }
 \vspace{-0.11in}
 \caption{\small Test NLL vs. performance of transductive inference with pair-wise distance metric. NLL is computed just before taking the initial transductive step.}

 \vspace{-0.20in}
 \label{tbl:ablation_ensemble}
\end{wraptable}

% \begin{table*}[t!]
% \centering
% \footnotesize
% \vspace{-0.05in}
% \caption{Ablation study on transduction with self-ensembling on miniImageNet.}
% \vspace*{0.03in}
%  \begin{tabular}{ c c | c c c c }

%  \hline
% Data & Model &
% \multicolumn{4}{c}{Transductive} \\
% Perturb & Perturb & NLL & 1-shot & NLL & 5-shot \\
%  \hline
%  \hline
% $\xmark$ & $\xmark$ &
% 1.1137(0.0117)/0.9841(0.0137) &
% 71.95\tiny$\pm$0.81 & 
% 0.8190(0.0079)/0.6324(0.0090) &
% 81.06\tiny$\pm$0.51 \\

% $\cmark$ & $\xmark$ &
% 1.0852(0.0152)/0.9885(0.0143) &
% 73.93\tiny$\pm$0.85 &
% 0.6824(0.0087)/0.6194(0.0084) &
% 81.93\tiny$\pm$0.49 \\

% $\xmark$ & $\cmark$ &
% 1.0355(0.0164)/1.0028(0.0162) &
% 74.07\tiny$\pm$0.85 &
% 0.6017(0.0087)/0.5938(0.0095) &
%  82.62\tiny$\pm$0.47 \\

% $\cmark$ & $\cmark$ &
% 1.0946(0.0173)/1.0596(0.0184) &
% \textbf{74.73\tiny$\pm$0.86} &
% 0.6026(0.0092)/0.5671(0.0096) &
% \textbf{83.36\tiny$\pm$0.45} \\
% \hline
%  \end{tabular}
%  \vspace{-0.15in}
%  \label{tbl:ablation_ensemble}
% \end{table*}
\paragraph{Effect of the model / data perturbation}
In Table~\ref{tbl:ablation_ensemble}, We analyze the contribution of each type of uncertainty to the reliability of confidence. We observe that the performance of transductive inference improves as we add in each type of uncertainties. We use negative log-likelihood (NLL) as the quality measure for the confidence scores: the lower the NLL, the closer the confidence scores to the target label. We observe that both types of uncertainties are helpful in improving the reliability of the output confidence.

% \vspace{-0.05in}
\begin{wrapfigure}{t}{0.5\textwidth}
    \vspace{-0.23in}
    \centering
    \subfigure{\includegraphics[width=0.51\linewidth]{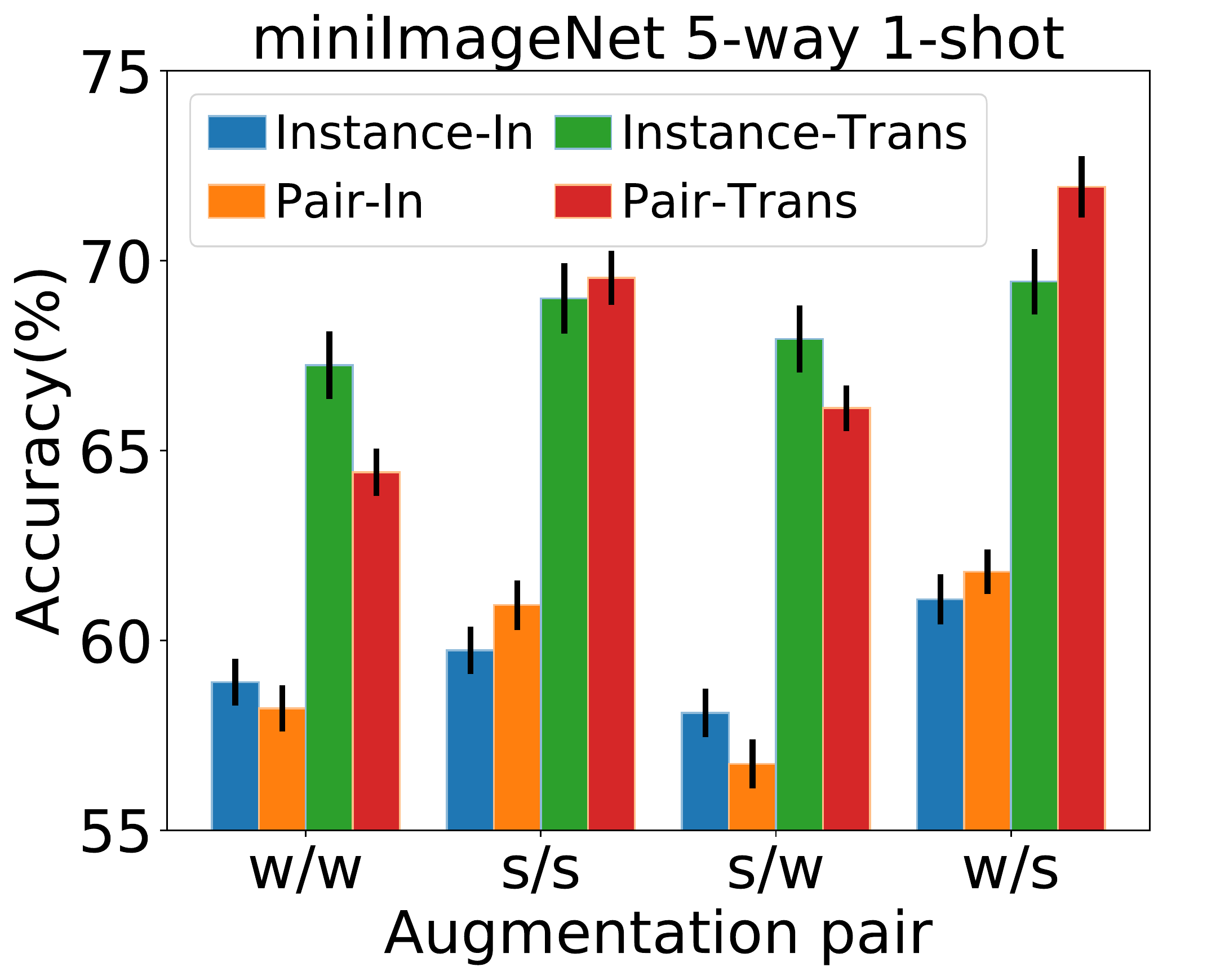}\label{aug_in}}
    \hspace{-0.1in}
    \subfigure{\includegraphics[width=0.49\linewidth]{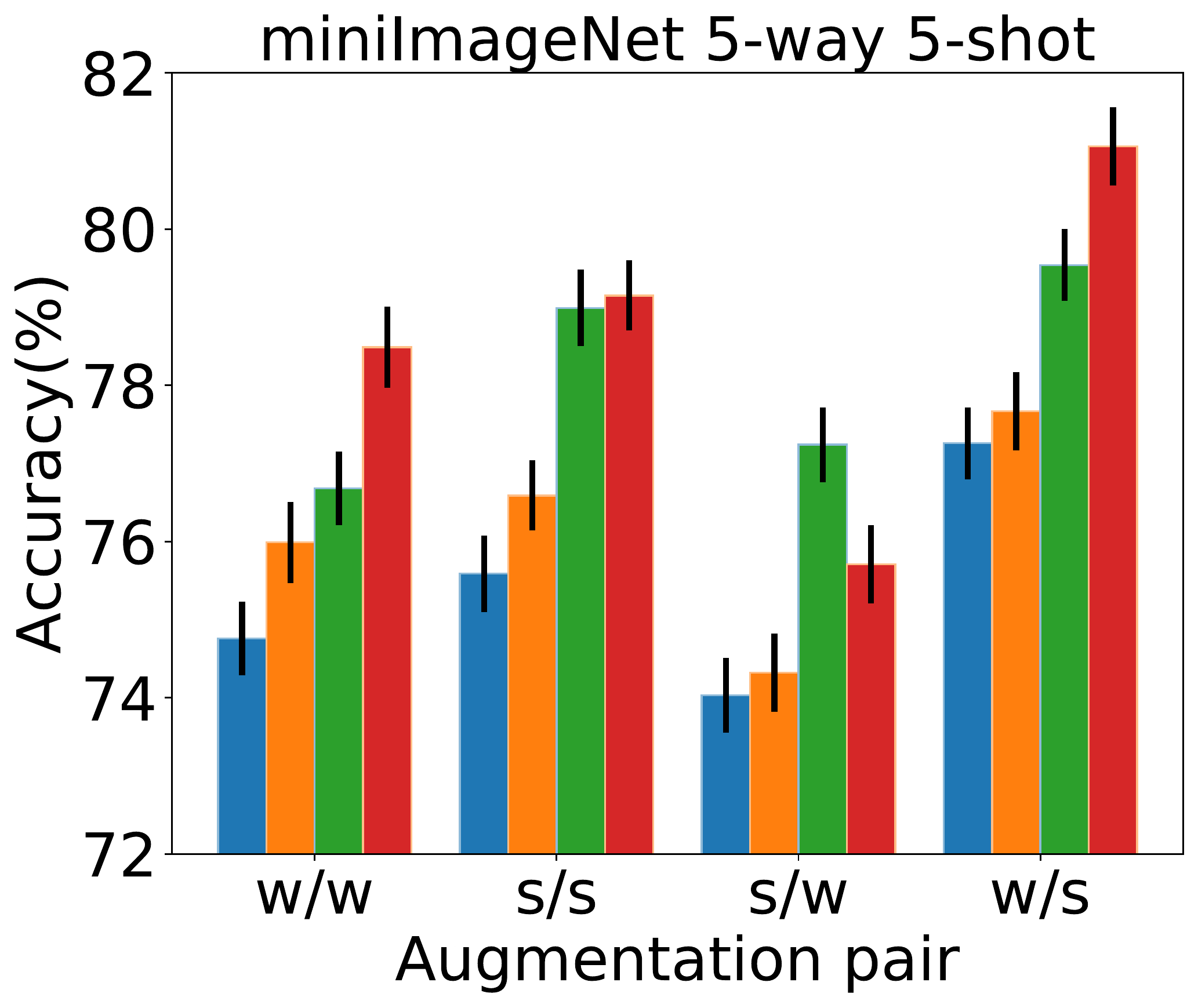}\label{aug_trans}}
    \vspace{-0.15in}
    \caption{\small Test accuracy with various augmentation pair. "w" and "s" denote "weak" and "strong", respectively. Detailed results with confidence interval can be found in Section~\ref{app:D} of the appendix.}\label{fig:aug}
    \vspace{-0.20in}
\end{wrapfigure}
\paragraph{Effect of the perturbation of query examples}
We next analyze the effect of our augmentation strategy. We see from Table~\ref{fig:aug} that applying weak augmentations (horizontal flipping and shifting) to the examples in the support set while applying strong augmentations (horizontal flipping, shifting, RandAugment~\cite{randaugment} and CutOut~\cite{cutout}) to the examples in the query set (weak-strong pair, i.e. w/s) outperforms other possible combinations of weak and strong augmentations. This result is reasonable since the class prototypes should remain as a stable target for stable training, while for query examples it may be beneficial to make them as diverse as possible, for the meta-learned confidence to account for various uncertainties with transductive inference.
%strong-strong pair achieves better result than weak-weak pair, where strong augmentation gives us more various training images and helps us reduce data cluster. 
%In fourth row, our proposed strategy obtains superior performance in both inductive and transductive settings. 
%However, strong-weak pair degrades performance rather than weak-weak and strong-strong pairs. If we apply strong augmentation to support set, the variance of prototypes will be large and make the training unstable. In contrast, our proposed setting, low variance for the prototypes and high variance for the query samples allow the model to be train stably and reduce variance of data according to data transformation.

\begin{wraptable}{t}{0.60\textwidth}
\begin{center}
	\vspace{-0.20in}
	\resizebox{0.6\textwidth}{!}{
		\small
		\begin{tabular}{cccc}
			\toprule
			Model &
            \multirow{2}{*}{Pooling} &
            \multicolumn{2}{c}{miniImageNet} \\
            (DC + ) &  & 1-shot & 5-shot \\
             \hline
             \hline
            Instance-wise metric ($d_\phi^I$) & 
            GAP &
            64.99\tiny$\pm$0.63&
            81.22\tiny$\pm$0.44 \\
            Instance-wise metric ($d_\phi^I$) & 
            None &
            \textbf{65.34\tiny$\pm$0.63} &
            \textbf{82.15\tiny$\pm$0.45} \\
             \hline
            Pair-wise metric ($d_\phi^P$) &
            GAP &
            62.66\tiny$\pm$0.62 &
            80.22\tiny$\pm$0.47 \\
            Pair-wise metric ($d_\phi^P$) &
            None &
            \textbf{64.49\tiny$\pm$0.64} &
            \textbf{81.63\tiny$\pm$0.44} \\
			\bottomrule	
		\end{tabular}
	}
	\vspace{-0.1in}
	\caption{\small The inductive inference performance with various dimension-wise classification methods.}\label{tbl:ablation_DC}
	\vspace{-0.15in}
\end{center}
\end{wraptable}
% \vspace{-0.05in}
\paragraph{Effect of the across-dimension consistency}
Lastly, we compare the effect of our consistency regularization across embedding dimensions with existing dense classification (DC) methods. In Table~\ref{tbl:ablation_DC}, we see that embedding without global average pooling (GAP) outperforms the model with GAP, demonstrating the effectiveness of dense matching of spatial features that gets rid of unnecessary bottlenecks. Also, unlike the existing DC which train pixel-wise classifiers during training and instance-wisely predict at test time, our method has a consistent framework as it has an additional instance-wise loss term (Eq.~\eqref{eq:inst_loss}) that is used both at training and test time.
%This performance gain is due to two reasons. First, our model computes the distance using dense matching of spatial features, while existing models use global average pooling before computing distances which results in the loss of spatial information. 

    \vspace{-0.05in}
\section{Conclusion}
\vspace{-0.05in}
Using unlabeled data for few-shot learning, either test instances themselves (transductive) or others (semi-supervised) could help with predictions. Yet, they should be assigned correct confidence scores for optimal performance gains. In this work, we proposed to tackle them by meta-learning confidence scores, such that the prototypes updated with meta-learend scores optimize for the transductive inference performance. Specifically, we proposed to \emph{meta-learn} the parameter of the length-scaling function, such that the proper \emph{distance metric} for the confidence scores can be automatically determined. We also consider model and data-level uncertainties for unseen examples, for more robust confidence estimation. Moreover, to enhance the quality of confidence scores, we suggest a consistency regularization for data and embedding, which allows for consistent prediction under various perturbations. We experimentally validate our transductive inference model on four benchmark datasets and obtain state-of-the-art performances on both transductive and semi-supervised few-shot classification tasks. Further ablation studies confirm the effectiveness of each component.

\section*{Broader Impact}
In real world scenarios, we may not have large amount of labeled data to train an accurate and reliable model on the target task, but should nevertheless obtain desired level of accuracy. To learn an accurate prediction model under such data-scarce scenarios, we may further exploit the unlabeled data given at test time (transductive inference), or extra unlabeled data during training (semi-supervised learning). Our model is especially helpful when using such unlabeled data to aid the learning with scarce data, as it is able to output accurate confidence scores for the unlabeled examples such that they help with the transductive inference or semi-supervised learning. Such low-resource learning can greatly reduce either the training time (for transductive inference) or human labeling cost (for semi-supervised learning) since we only need a few training data points to train a classifier that obtains the desirable level of accuracy.  
	\bibliographystyle{abbrv}
	\bibliography{refs}

\begin{thebibliography}{10}

\bibitem{SCA}
A.~Antoniou and A.~J. Storkey.
\newblock Learning to learn by self-critique.
\newblock In {\em NeurIPS}, pages 9936--9946, 2019.

\bibitem{bertinetto2018metalearning}
L.~Bertinetto, J.~F. Henriques, P.~Torr, and A.~Vedaldi.
\newblock Meta-learning with differentiable closed-form solvers.
\newblock In {\em ICLR}, 2019.

\bibitem{randaugment}
E.~D. Cubuk, B.~Zoph, J.~Shlens, and Q.~V. Le.
\newblock Randaugment: Practical data augmentation with no separate search.
\newblock {\em arXiv preprint arXiv:1909.13719}, 2019.

\bibitem{das2019two}
D.~Das and C.~G. Lee.
\newblock A two-stage approach to few-shot learning for image recognition.
\newblock {\em IEEE Transactions on Image Processing}, 2019.

\bibitem{cutout}
T.~DeVries and G.~W. Taylor.
\newblock Improved regularization of convolutional neural networks with cutout.
\newblock {\em arXiv preprint arXiv:1708.04552}, 2017.

\bibitem{baseline}
G.~S. Dhillon, P.~Chaudhari, A.~Ravichandran, and S.~Soatto.
\newblock A baseline for few-shot image classification.
\newblock In {\em ICLR}, 2020.

\bibitem{finn2017model}
C.~Finn, P.~Abbeel, and S.~Levine.
\newblock Model-agnostic meta-learning for fast adaptation of deep networks.
\newblock In {\em ICML}, pages 1126--1135. JMLR. org, 2017.

\bibitem{dropout_as_bayesian}
Y.~{Gal} and Z.~{Ghahramani}.
\newblock {Dropout as a Bayesian Approximation: Representing Model Uncertainty
  in Deep Learning}.
\newblock In {\em ICML}, 2016.

\bibitem{gordon2018meta}
J.~Gordon, J.~Bronskill, M.~Bauer, S.~Nowozin, and R.~E. Turner.
\newblock Meta-learning probabilistic inference for prediction.
\newblock In {\em ICLR}, 2018.

\bibitem{grant2018recasting}
E.~Grant, C.~Finn, S.~Levine, T.~Darrell, and T.~Griffiths.
\newblock Recasting gradient-based meta-learning as hierarchical bayes.
\newblock In {\em ICLR}, 2018.

\bibitem{hou2019cross}
R.~Hou, H.~Chang, M.~Bingpeng, S.~Shan, and X.~Chen.
\newblock Cross attention network for few-shot classification.
\newblock In {\em NeurIPS}, pages 4005--4016, 2019.

\bibitem{SIB}
S.~X. Hu, P.~G. Moreno, Y.~Xiao, X.~Shen, G.~Obozinski, N.~D. Lawrence, and
  A.~Damianou.
\newblock Empirical bayes transductive meta-learning with synthetic gradients.
\newblock In {\em ICLR}, 2020.

\bibitem{kim2019edge}
J.~Kim, T.~Kim, S.~Kim, and C.~D. Yoo.
\newblock Edge-labeling graph neural network for few-shot learning.
\newblock In {\em CVPR}, pages 11--20, 2019.

\bibitem{temporal}
S.~Laine and T.~Aila.
\newblock Temporal ensembling for semi-supervised learning.
\newblock In {\em ICLR}, 2017.

\bibitem{lee2019meta}
K.~Lee, S.~Maji, A.~Ravichandran, and S.~Soatto.
\newblock Meta-learning with differentiable convex optimization.
\newblock In {\em CVPR}, pages 10657--10665, 2019.

\bibitem{lee2018gradient}
Y.~Lee and S.~Choi.
\newblock Gradient-based meta-learning with learned layerwise metric and
  subspace.
\newblock In {\em ICML}, pages 2933--2942, 2018.

\bibitem{li2019learning}
X.~Li, Q.~Sun, Y.~Liu, Q.~Zhou, S.~Zheng, T.-S. Chua, and B.~Schiele.
\newblock Learning to self-train for semi-supervised few-shot classification.
\newblock In {\em NeurIPS}, pages 10276--10286, 2019.

\bibitem{l2ST}
X.~Li, Q.~Sun, Y.~Liu, Q.~Zhou, S.~Zheng, T.-S. Chua, and B.~Schiele.
\newblock Learning to self-train for semi-supervised few-shot classification.
\newblock In {\em NeurIPS}, pages 10276--10286, 2019.

\bibitem{lifchitz2019dense}
Y.~Lifchitz, Y.~Avrithis, S.~Picard, and A.~Bursuc.
\newblock Dense classification and implanting for few-shot learning.
\newblock In {\em CVPR}, pages 9258--9267, 2019.

\bibitem{liu2018learning}
Y.~Liu, J.~Lee, M.~Park, S.~Kim, E.~Yang, S.~J. Hwang, and Y.~Yang.
\newblock Learning to propagate labels: Transductive propagation network for
  few-shot learning.
\newblock In {\em ICLR}, 2018.

\bibitem{l2b}
D.~Na, H.~B. Lee, H.~Lee, S.~Kim, M.~Park, E.~Yang, and S.~J. Hwang.
\newblock Learning to balance: Bayesian meta-learning for imbalanced and
  out-of-distribution tasks.
\newblock In {\em ICLR}, 2020.

\bibitem{oreshkin2018tadam}
B.~Oreshkin, P.~R. L{\'o}pez, and A.~Lacoste.
\newblock Tadam: Task dependent adaptive metric for improved few-shot learning.
\newblock In {\em NeurIPS}, pages 721--731, 2018.

\bibitem{trans_episodic}
L.~Qiao, Y.~Shi, J.~Li, Y.~Wang, T.~Huang, and Y.~Tian.
\newblock Transductive episodic-wise adaptive metric for few-shot learning.
\newblock In {\em ICCV}, pages 3603--3612, 2019.

\bibitem{opt_as_model}
S.~Ravi and H.~Larochelle.
\newblock Optimization as a model for few-shot learning.
\newblock In {\em ICLR}, 2017.

\bibitem{ren2018meta}
M.~Ren, E.~Triantafillou, S.~Ravi, J.~Snell, K.~Swersky, J.~B. Tenenbaum,
  H.~Larochelle, and R.~S. Zemel.
\newblock Meta-learning for semi-supervised few-shot classification.
\newblock In {\em ICLR}, 2018.

\bibitem{cnaps}
J.~Requeima, J.~Gordon, J.~Bronskill, S.~Nowozin, and R.~E. Turner.
\newblock Fast and flexible multi-task classification using conditional neural
  adaptive processes.
\newblock In {\em NeurIPS}, pages 7957--7968, 2019.

\bibitem{russakovsky2015imagenet}
O.~Russakovsky, J.~Deng, H.~Su, J.~Krause, S.~Satheesh, S.~Ma, Z.~Huang,
  A.~Karpathy, A.~Khosla, M.~Bernstein, et~al.
\newblock Imagenet large scale visual recognition challenge.
\newblock {\em ICJV}, 115(3):211--252, 2015.

\bibitem{leo}
A.~A. Rusu, D.~Rao, J.~Sygnowski, O.~Vinyals, R.~Pascanu, S.~Osindero, and
  R.~Hadsell.
\newblock Meta-learning with latent embedding optimization.
\newblock In {\em ICLR}, 2019.

\bibitem{snell2017prototypical}
J.~Snell, K.~Swersky, and R.~Zemel.
\newblock Prototypical networks for few-shot learning.
\newblock In {\em NeurIPS}, pages 4077--4087, 2017.

\bibitem{fixmatch}
K.~Sohn, D.~Berthelot, C.-L. Li, Z.~Zhang, N.~Carlini, E.~D. Cubuk, A.~Kurakin,
  H.~Zhang, and C.~Raffel.
\newblock Fixmatch: Simplifying semi-supervised learning with consistency and
  confidence.
\newblock {\em arXiv preprint arXiv:2001.07685}, 2020.

\bibitem{MTL}
Q.~Sun, Y.~Liu, T.-S. Chua, and B.~Schiele.
\newblock Meta-transfer learning for few-shot learning.
\newblock In {\em CVPR}, pages 403--412, 2019.

\bibitem{sung2018learning}
F.~Sung, Y.~Yang, L.~Zhang, T.~Xiang, P.~H. Torr, and T.~M. Hospedales.
\newblock Learning to compare: Relation network for few-shot learning.
\newblock In {\em CVPR}, pages 1199--1208, 2018.

\bibitem{mean_teacher}
A.~Tarvainen and H.~Valpola.
\newblock Mean teachers are better role models: Weight-averaged consistency
  targets improve semi-supervised deep learning results.
\newblock In {\em NeurIPS}, pages 1195--1204, 2017.

\bibitem{thrun98}
S.~Thrun and L.~Pratt.
\newblock {\em Learning to Learn}.
\newblock Kluwer Academic Publishers, Norwell, MA, USA, 1998.

\bibitem{triantafillou2019meta}
E.~Triantafillou, T.~Zhu, V.~Dumoulin, P.~Lamblin, K.~Xu, R.~Goroshin,
  C.~Gelada, K.~Swersky, P.-A. Manzagol, and H.~Larochelle.
\newblock Meta-dataset: A dataset of datasets for learning to learn from few
  examples.
\newblock {\em arXiv preprint arXiv:1903.03096}, 2019.

\bibitem{vapnik98}
V.~N. Vapnik.
\newblock {\em Statistical Learning Theory}.
\newblock Wiley, 1998.

\bibitem{veit2016residual}
A.~Veit, M.~J. Wilber, and S.~Belongie.
\newblock Residual networks behave like ensembles of relatively shallow
  networks.
\newblock In {\em NeurIPS}, pages 550--558, 2016.

\bibitem{vinyals2016matching}
O.~Vinyals, C.~Blundell, T.~Lillicrap, D.~Wierstra, et~al.
\newblock Matching networks for one shot learning.
\newblock In {\em NeurIPS}, pages 3630--3638, 2016.

\bibitem{blockdrop}
Z.~Wu, T.~Nagarajan, A.~Kumar, S.~Rennie, L.~S. Davis, K.~Grauman, and
  R.~Feris.
\newblock {BlockDrop: Dynamic Inference Paths in Residual Networks}.
\newblock In {\em CVPR}, 2018.

\bibitem{DPGN}
L.~Yang, L.~Li, Z.~Zhang, E.~Zhou, Y.~Liu, et~al.
\newblock Dpgn: Distribution propagation graph network for few-shot learning.
\newblock In {\em CVPR}, 2020.

\bibitem{yoon2019tapnet}
S.~W. Yoon, J.~Seo, and J.~Moon.
\newblock Tapnet: Neural network augmented with task-adaptive projection for
  few-shot learning.
\newblock {\em arXiv preprint arXiv:1905.06549}, 2019.

\bibitem{cavia}
L.~Zintgraf, K.~Shiarli, V.~Kurin, K.~Hofmann, and S.~Whiteson.
\newblock Fast context adaptation via meta-learning.
\newblock In {\em ICML}, 2019.

\end{thebibliography}

    \newpage
    \maketitle
    \begin{appendices}
    \section{Experimental Setup}\label{app:A}

\subsection{Datasets}\label{app:A.1}
We validate our method on four benckmark datasets for few-shot classification.

\textbf{1) miniImageNet.} This dataset~\cite{vinyals2016matching} consists of a subset of $100$ classes sampled from the ImageNet dataset~\cite{russakovsky2015imagenet}. Each class has $600$ images, resized to $84\times84$ pixels. We use the split of $64/16/20$ for training/validation/test.

\textbf{2) tieredImageNet.} This dataset~\cite{ren2018meta} is another subset of ImageNet, that consists of $779,165$ images of $84\times{84}$ pixels collected from $608$ classes. The task is to generalize the few-shot classifier over $34$ different superclasses. Thus the entire dataset is split into $20/6/8$ superclasses for training/validation/test, where each superclass contains $351$, $97$, and $160$ low-level classes, respectively.  

\textbf{3) CIFAR-FS.} This dataset~\cite{bertinetto2018metalearning} is a variant of CIFAR-100 dataset used for few-shot classification, which contains $100$ classes that describe general object categories. For each class, there are $600$ images of $32\times{32}$ pixels. The dataset is split into $64/16/20$ classes for training/validation/test.

\textbf{4) FC100.} This is another few-shot classification dataset~\cite{oreshkin2018tadam} compiled by reorganizing the CIFAR-100 dataset. The task for this dataset is to generalize across $20$ superclasses, as done with the tieredImageNet dataset. The superclasses are divided into $12/4/4$ classes for training/validation/test, each of which contains $60/20/20$ low-level classes, respectively. 

\subsection{Network architectures}\label{app:A.2}
We consider ResNet-12 backbone and conventional 4-block convolutional networks with 64-64-64-64 (ConvNet-64) or 64-96-128-256 (ConvNet-256) channels for each layer. We implement the metric scaling function as a single convolutional block followed by two fully-connected layers (FC-layers). The convolutional block consists of 3x3 convolution, batch normalization, ReLU activation and 2x2 max pooling. The first FC-layer is followed by batch normalization and ReLU activation, whereas the last FC-layer followed by sigmoid function to ensure non-negativity. Finally, in order to balance the effect of the scaling and normalized distance on confidence, we apply scaling ($exp(\alpha)$) and shifting ($exp(\beta)$) to the output of the sigmoid function, where $\alpha$ and $\beta$ are initialized to 0.

\subsection{Hyperparameters}\label{app:A.3}
We apply dropout to each layer with the ratio of $0.1$. We use SGD optimizer with the Nesterov momentum of $0.9$ and set the weight decay to $0.0005$. Following Snell \textit{et al.}~\cite{snell2017prototypical}, we use higher way ($15$-way) classification for training and $5$-way for test. The number of query examples for each class is set to $8$ for training and $15$ for test. For \textbf{miniImageNet}, \textbf{CIFAR-FS} and \textbf{FC100}, we set the initial learning rate to $0.1$ and cut it to $0.006$ and $0.0012$ at $25,000$ and $35,000$ episodes, respectively. For \textbf{tieredImageNet}, we set the initial learning rate to $0.1$ and decay it by a factor of 10 at every $20,000$ episode until convergence.

\section{Settings for Semi-Supervised Few-shot Classification}\label{app:B}
We split both miniImageNet and tieredImageNet into labeled and unlabeled sets, following previous works~\cite{l2ST, ren2018meta}. Before we train the model with semi-supervised learning, we pre-train the model with conventional supervised manner (e.g. 64-way classification for miniImageNet). At the training phase, we additionally use 15 instances for each class. At test phase, we use 30 and 50 unlabeled instances for each class on 1-shot and 5-shot task, respectively, following Li \textit{et al.}~\cite{l2ST}. For fair comparison with masked soft $k$-means of Ren \textit{et al.}~\cite{ren2018meta}, we use single update step with unlabeled set for both training and testing.

    \section{Detailed Explanation of Meta-Confidence Transduction}\label{app:C}

\subsection{Design choices}
\paragraph{Model perturbation} To generate the model uncertainty, we drop the last residual block in residual network (ResNet). As discussed in Veit \textit{et al.}~\cite{veit2016residual}, dropping single upper block in ResNet doesn't significantly affect model performance. Furthermore, we empirically found that block drop allows us to obtain the model with less dependency rather than dropout.

\paragraph{Data perturbation}
There can be various choices of augmentation method to perturb the data. However, we found that large transformation from the raw image can degrade classification accuracy at inference, causing large information loss on few data regime. Thus, we choose horizontal flip augmentation, which can perturb the data without losing information, to obtain perturbed confidences in training and testing consistently.

\paragraph{Optimization}
The reason we optimize only a single full-path is as follows. First, since we randomly apply horizontal flipping to the whole data in each episode, perturbed spaces with flipped images are optimized through the sequence of episodes. Secondly, as drop-path is one of the ensemble path of full-path, it is jointly optimized with full-path~\cite{veit2016residual}.
\begin{algorithm}[h!]
  \caption{Meta-Confidence Transduction (MCT)}
    \begin{flushleft}
        \textbf{Require:} The number of classes $C$, and the number of transduction steps $T$.
        \\
        \textbf{Require:} The set of support examples $\mathcal{S}_c$, for each class $c=1,\dots,C$.
        \\
        \textbf{Require:} The set of all query examples $\mathcal{Q}_x$.
        \\
        \textbf{Require:} Full-path embedding function $f_\theta$ and block-dropped embedding function $f^D_\theta$.
        \\
        \textbf{Require:} Flip augmentation $\text{Aug}(\cdot)$ and define $f_{\theta}^{A}$ as $f_{\theta}(\text{Aug}(\cdot))$.
        \\
        \textbf{Require:} Embedding function set $F=\{f_{\theta}, 
        f^{D}_{\theta}, 
        f^{A}_{\theta}, 
        f^{A,D}_{\theta}
        \}$
        \\
        \textbf{Output:} Confidence score $q_c^{T}(\btx)$ obtained after $T$ transduction steps, for all $c=1,\dots,C$ and $\btx \in \mathcal{Q}_x$.
        \\
    \end{flushleft}
  \begin{algorithmic}[1]
  
    \For{$c = 1,\dots, C$}
        \State $P_{0,c}^{h_{\theta}} \gets \frac{1}{|\mathcal{S}_c|} \sum_{\bx \in \mathcal{S}_c} h_\theta(\bx)$ for all $h_{\theta} \in F$ \Comment{Compute initial prototype for each space}
    \EndFor
    \For{$t=0,\dots, T$}
        \State $q^{(t)}_{c}(\btx) \leftarrow 0$
        \Comment{Initialize confidence score}
        \For{$c = 1,\dots, C$}
            \For{$h_{\theta}$ in $F$}
                \State $\sigma_{t,c}^{h_{\theta}}(\btx) \gets \cfrac{\exp(-d_\phi(h_\theta(\btx),P_{t,c}^{h_\theta})}{\sum_{c'}^C\exp(-d_\phi(h_\theta(\btx),P_{t,c'}^{h_\theta})}$ for all $\btx \in \mathcal{Q}_x$ \Comment{Compute local confidence}
                \State $q^{(t)}_{c}(\btx) \leftarrow 
                q^{(t)}_{c}(\btx)+\cfrac{1}{|F|} \cdot \sigma_{t,c}^{h_{\theta}}(\btx)$ for all $\btx \in \mathcal{Q}_x$
                \Comment{Obtain ensemble confidence score}
            \EndFor
            \For{$h_{\theta}$ in $F$}
            \State $P_{t+1,c}^{h_{\theta}} \gets \cfrac{\sum_{\bx \in \mathcal{S}_c}1 \cdot h_\theta(\bx)+\sum_{\btx \in \mathcal{Q}_x}q_c^{(t)}(\btx) \cdot h_\theta(\btx)}{\sum_{\bx \in \mathcal{S}_c}1+\sum_{\btx \in \mathcal{Q}_x}q_c^{(t)}(\btx)}$\\                  \Comment{Update class $c$ prototype for each space}
            \EndFor
        \EndFor
    \EndFor
  \end{algorithmic}
  \label{algo:test_MCT}
\end{algorithm}
\subsection{Transductive inference}
As shown in Algorithm~\ref{algo:test_MCT}, we update the class prototypes by considering various types of uncertainties. Given an episode consisting of raw images, we generate another episode by flipping the original images. First, prototypes of full-path and drop-path are obtained by averaging embedding of support set. By using these prototypes, we compute the confidence scores for each space and class, respectively. With the ensemble confidence score obtained from various spaces and queries, we update prototypes of each space. Then, we repeatedly update the prototype $T$ times by using an averaged confidence. Finally, $q^{(T)}(\btx)$ is used for inference.

\begin{table*}[ht!]
\centering
\footnotesize
% \vspace*{0.03in}
 \begin{tabular}{ c  c  c  c  c  c  c  c }
 \hline
\multirow{2}{*}{Model}  & \multicolumn{2}{c}{Inductive} & \multicolumn{2}{c}{Transductive}\\
  & 1-shot & 5-shot & 1-shot & 5-shot\\
 \hline
 \hline
MCI (Pair) &
64.49\tiny$\pm$0.64 & 81.63\tiny$\pm$0.44 &
75.07\tiny$\pm$0.89 & 84.09\tiny$\pm$0.47 \\

MCT (Pair) &
\textbf{64.89\tiny$\pm$0.63} &
\textbf{82.48\tiny$\pm$0.42} &
\textbf{76.16\tiny$\pm$0.89} &
\textbf{85.22\tiny$\pm$0.42} \\
\hline
MCI (Instance) &
65.34\tiny$\pm$0.63 & 82.15\tiny$\pm$0.45 &
76.21\tiny$\pm$0.82 & 84.49\tiny$\pm$0.46 \\

MCT (Instance) &
\textbf{66.47\tiny$\pm$0.63} &
\textbf{83.29\tiny$\pm$0.42} &
\textbf{78.55\tiny$\pm$0.86} &
\textbf{86.03\tiny$\pm$0.42} \\
\hline
 \end{tabular}
 \caption{Effect of model and data perturbations.}
%  \vspace{-0.15in}
 \label{tbl:MCI_MCT}
\end{table*}
\vspace{-0.2in}
\subsection{Importance of model and data perturbations}
To further investigate the effect of the model and data perturbations on confidence, we report more results of MCI and MCT in Table~\ref{tbl:MCI_MCT}. MCI is the model that is trained with inductive manner, whereas MCT is transductively trained with model and data perturbations. We see that the MCT outperforms MCI in both inductive and transductive settings. It means that meta-learned confidence with perturbations allows us to obtain more reliable confidence, which is further helpful for transduction.
\vspace{+0.1in}
    \vspace{+0.3in}
\section{Detailed Result for Augmentation Strategy}\label{app:D}
% \vspace{-0.05in}
\begin{table*}[h!]
\centering
\footnotesize
% \vspace{-0.05in}
 \caption{Result of augmentation strategy on pair-wise distance metric.}
\vspace*{0.03in}
 \begin{tabular}{c c | c c c c }

 \hline
Support & Query & \multicolumn{2}{c}{Inductive} & \multicolumn{2}{c}{Transductive} \\
Aug & Aug & 1-shot & 5-shot & 1-shot & 5-shot \\
 \hline
 \hline
weak & weak &
58.21\tiny$\pm$0.61 & 75.99\tiny$\pm$0.52 &
64.43\tiny$\pm$0.62 & 78.49\tiny$\pm$0.50 \\

strong & strong &
60.93\tiny$\pm$0.65 & 76.59\tiny$\pm$0.45 &
69.55\tiny$\pm$0.71 & 79.15\tiny$\pm$0.52 \\

strong & weak   &
56.75\tiny$\pm$0.64 & 74.32\tiny$\pm$0.50 &
66.12\tiny$\pm$0.60 & 75.71\tiny$\pm$0.52 \\

weak & strong   &
\textbf{61.81\tiny$\pm$0.58} &
\textbf{77.67\tiny$\pm$0.50} &
\textbf{71.95\tiny$\pm$0.81} & 
\textbf{81.06\tiny$\pm$0.51} \\
\hline
 \end{tabular}
 \label{tbl:ablation_aug_PMS}
\end{table*}
%  \vspace{+3in}

\begin{table*}[h!]
\centering
\footnotesize
% \vspace{-0.05in}
 \caption{Result of augmentation strategy on instance-wise distance metric.}
\vspace*{0.03in}
 \begin{tabular}{c c | c c c c }

 \hline
Support & Query & \multicolumn{2}{c}{Inductive} & \multicolumn{2}{c}{Transductive} \\
Aug & Aug & 1-shot & 5-shot & 1-shot & 5-shot \\
 \hline
 \hline
weak & weak &
58.90\tiny$\pm$0.61 &
74.76\tiny$\pm$0.47 &
67.25\tiny$\pm$0.89 &
76.68\tiny$\pm$0.55	 \\

strong & strong &
59.74\tiny$\pm$0.62 &
75.59\tiny$\pm$0.49 &
69.01\tiny$\pm$0.93 &
78.99\tiny$\pm$0.53	 \\

strong & weak   &
58.09\tiny$\pm$0.64 &
74.03\tiny$\pm$0.48 &
67.94\tiny$\pm$0.88 & 
77.24\tiny$\pm$0.55 \\

weak & strong   &
\textbf{61.08\tiny$\pm$0.66} &
\textbf{77.26\tiny$\pm$0.46} &
\textbf{69.45\tiny$\pm$0.86} & 
\textbf{79.54\tiny$\pm$0.54} \\
\hline
 \end{tabular}
%  \vspace{-0.15in}
 \label{tbl:ablation_aug_IMS}
\end{table*}
    \newpage
\vspace{+1.65in}
\section{Qualitative Analysis}
\vspace{-0.2in}
\begin{figure}[h!]
\small
\centering
    \subfigure[\small Raw image, Full-path, Local conf]{\includegraphics[width=0.42\linewidth]{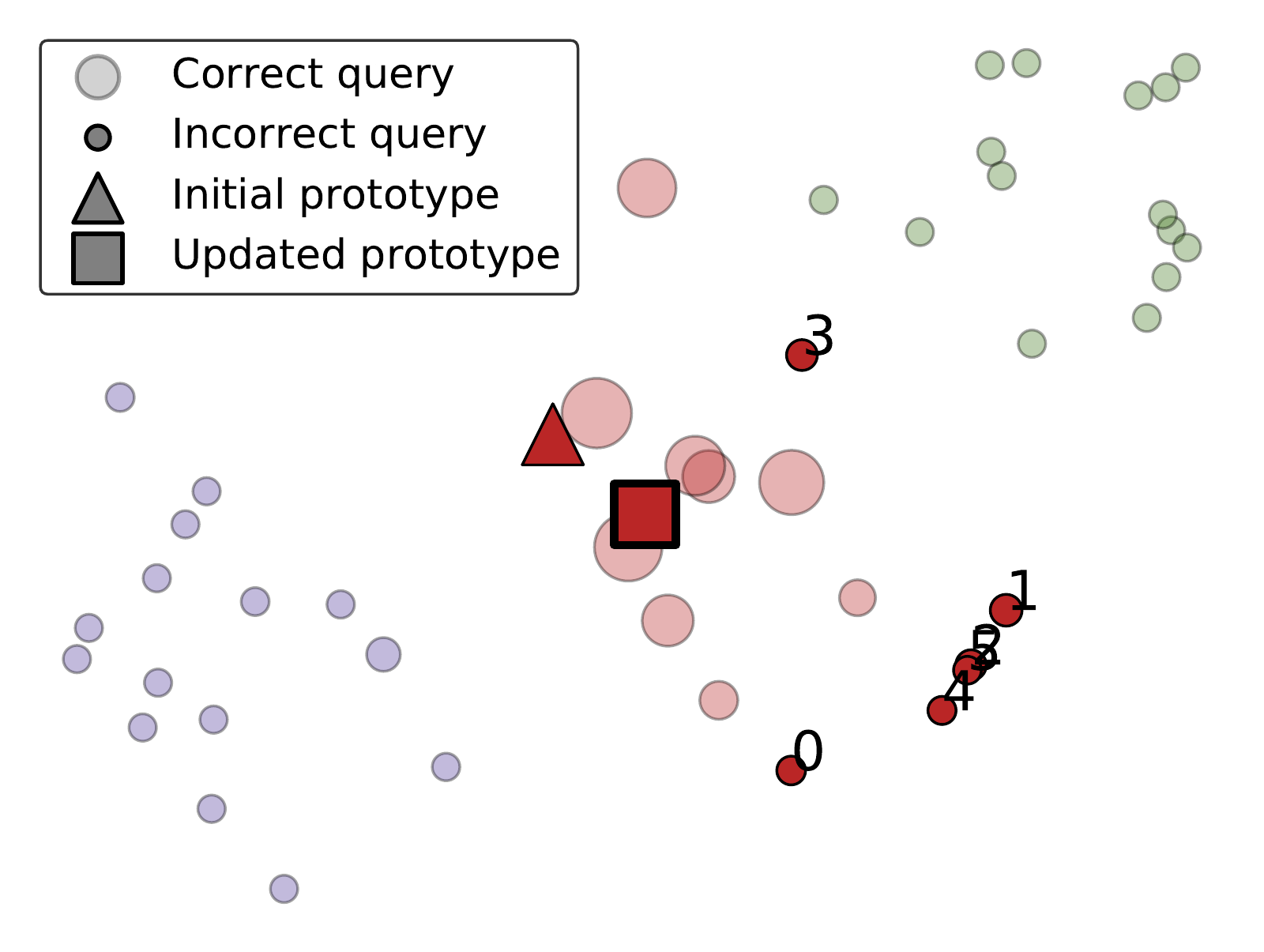}}
\hfil
    \subfigure[\small Raw image, Full-path, Ensemble conf]{\includegraphics[width=0.42\linewidth]{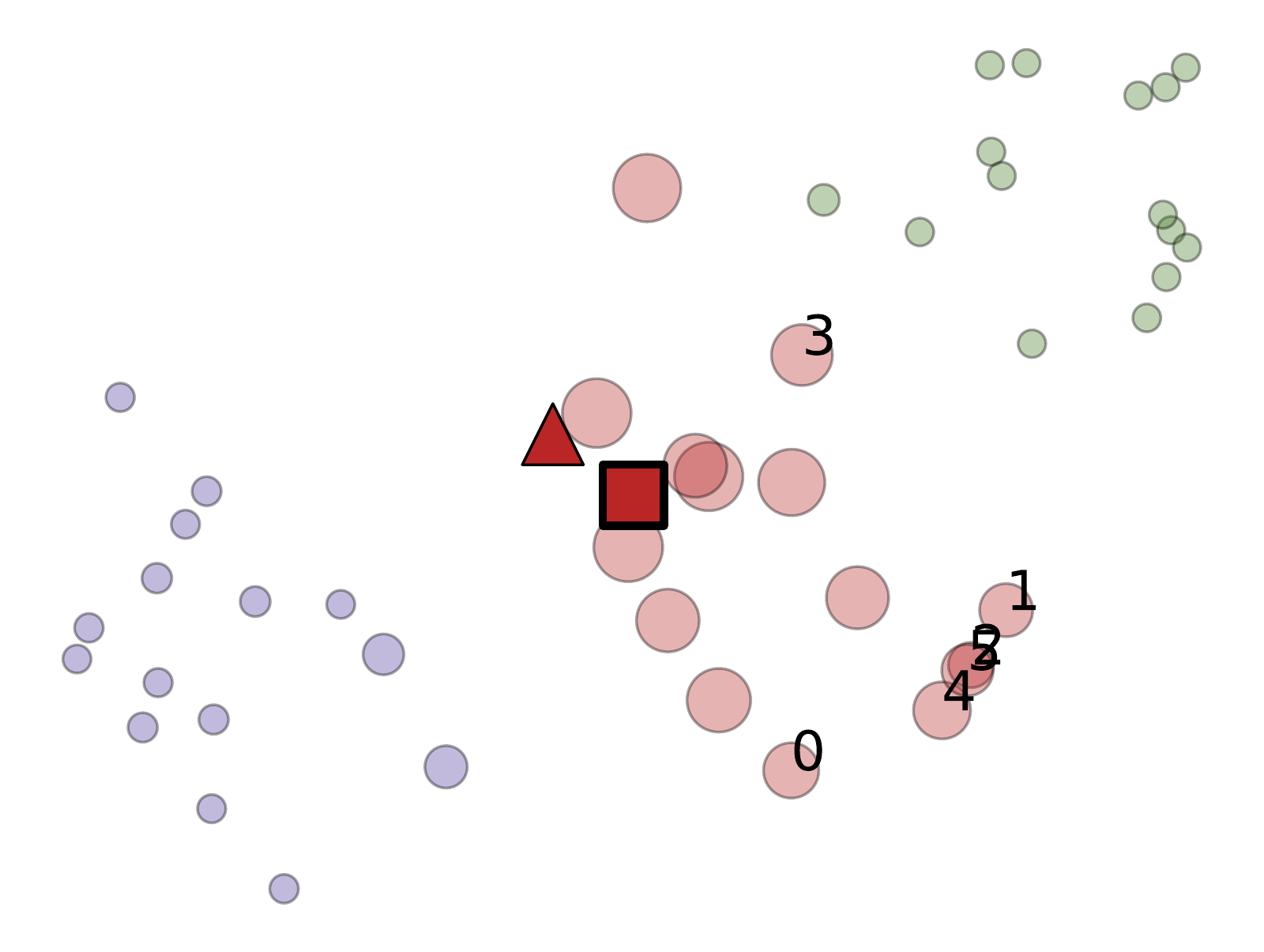}}
\vspace{-0.1in}
    \subfigure[\small Raw image, Drop-path, Local conf]{\includegraphics[width=0.42\linewidth]{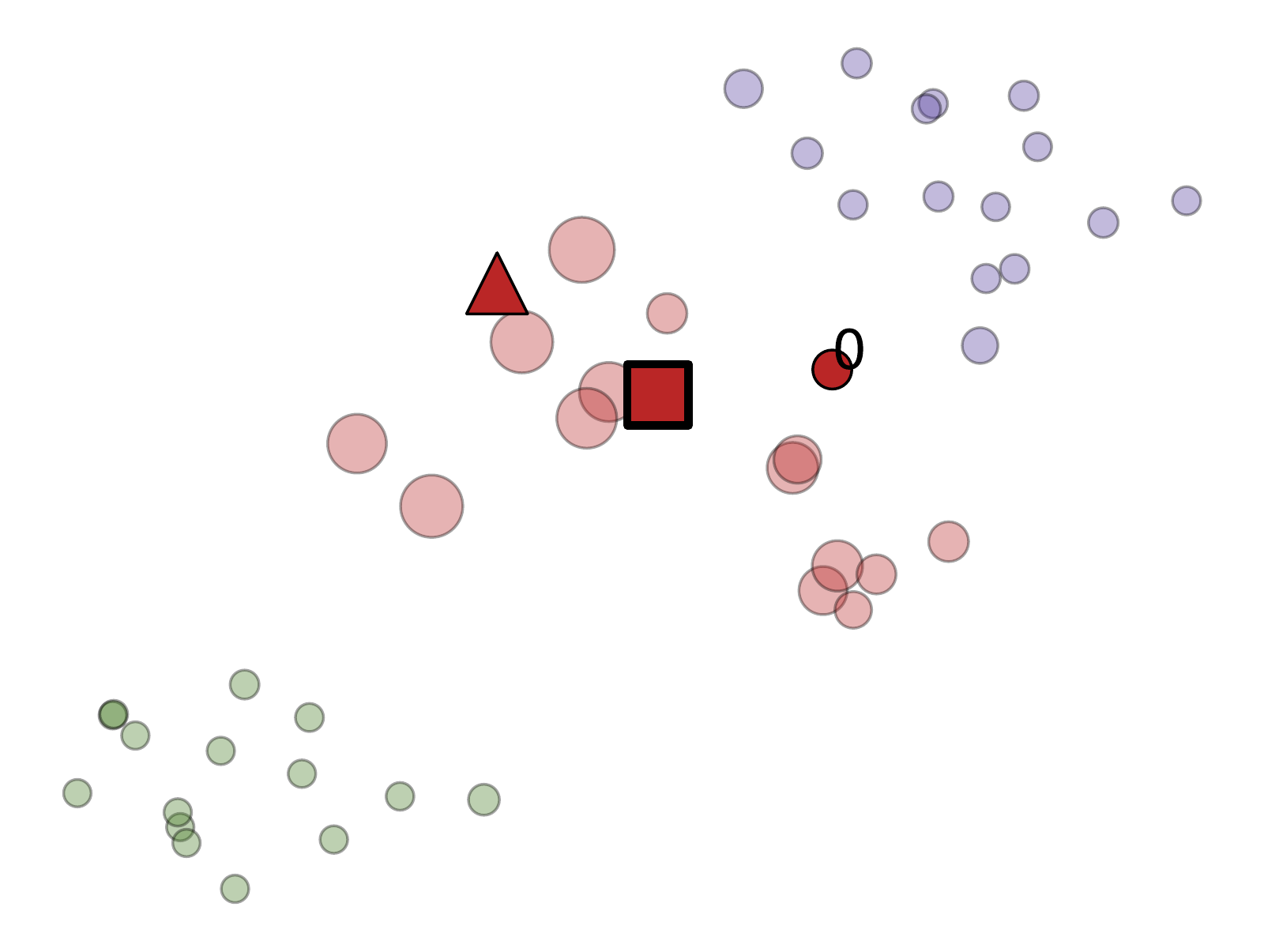}}
\hfil
    \subfigure[\small Raw image, Drop-path, Ensemble conf]{\includegraphics[width=0.42\linewidth]{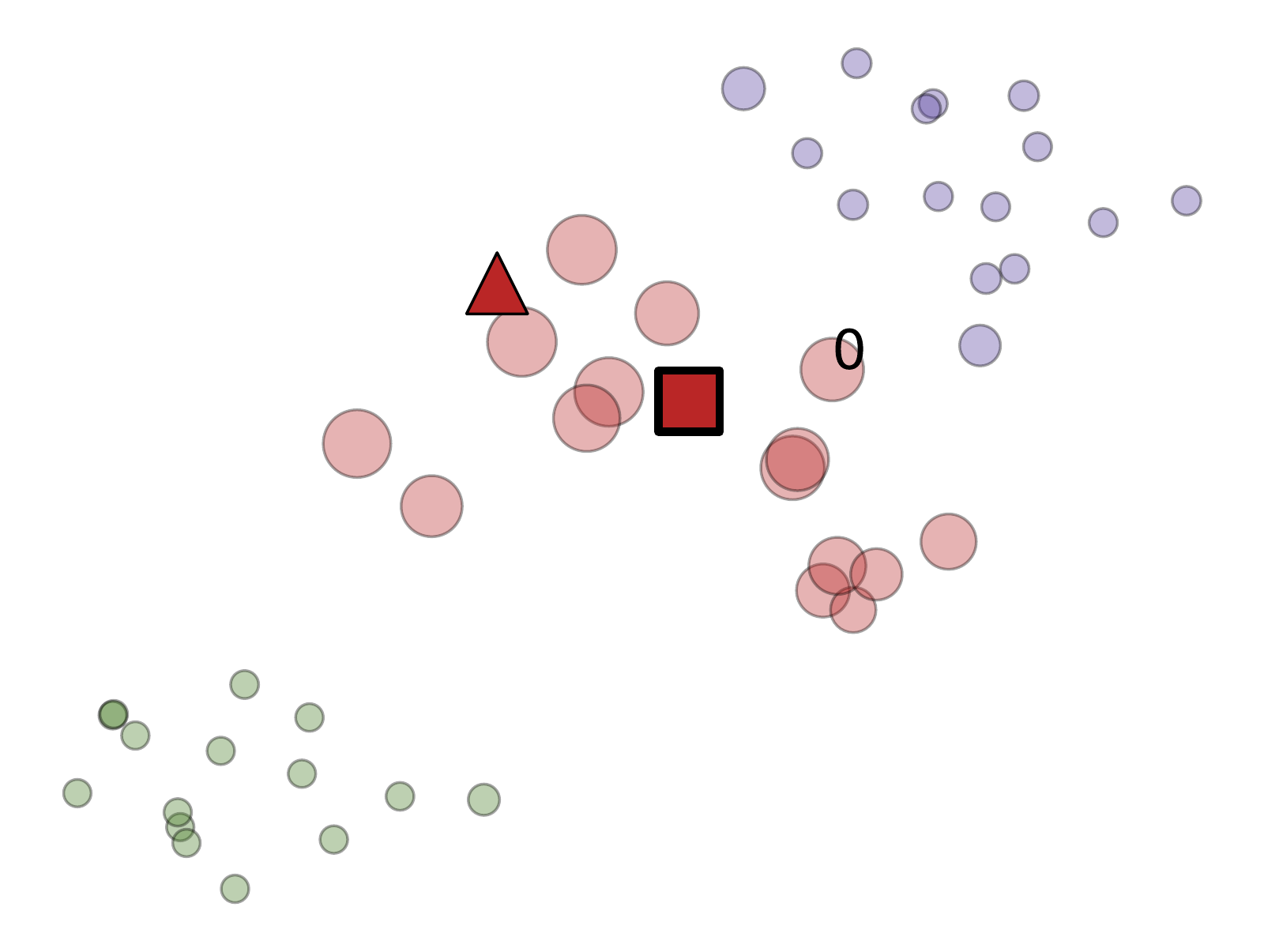}}
\vspace{-0.1in}
    \subfigure[\small Flipped image, Full-path, Local conf]{\includegraphics[width=0.42\linewidth]{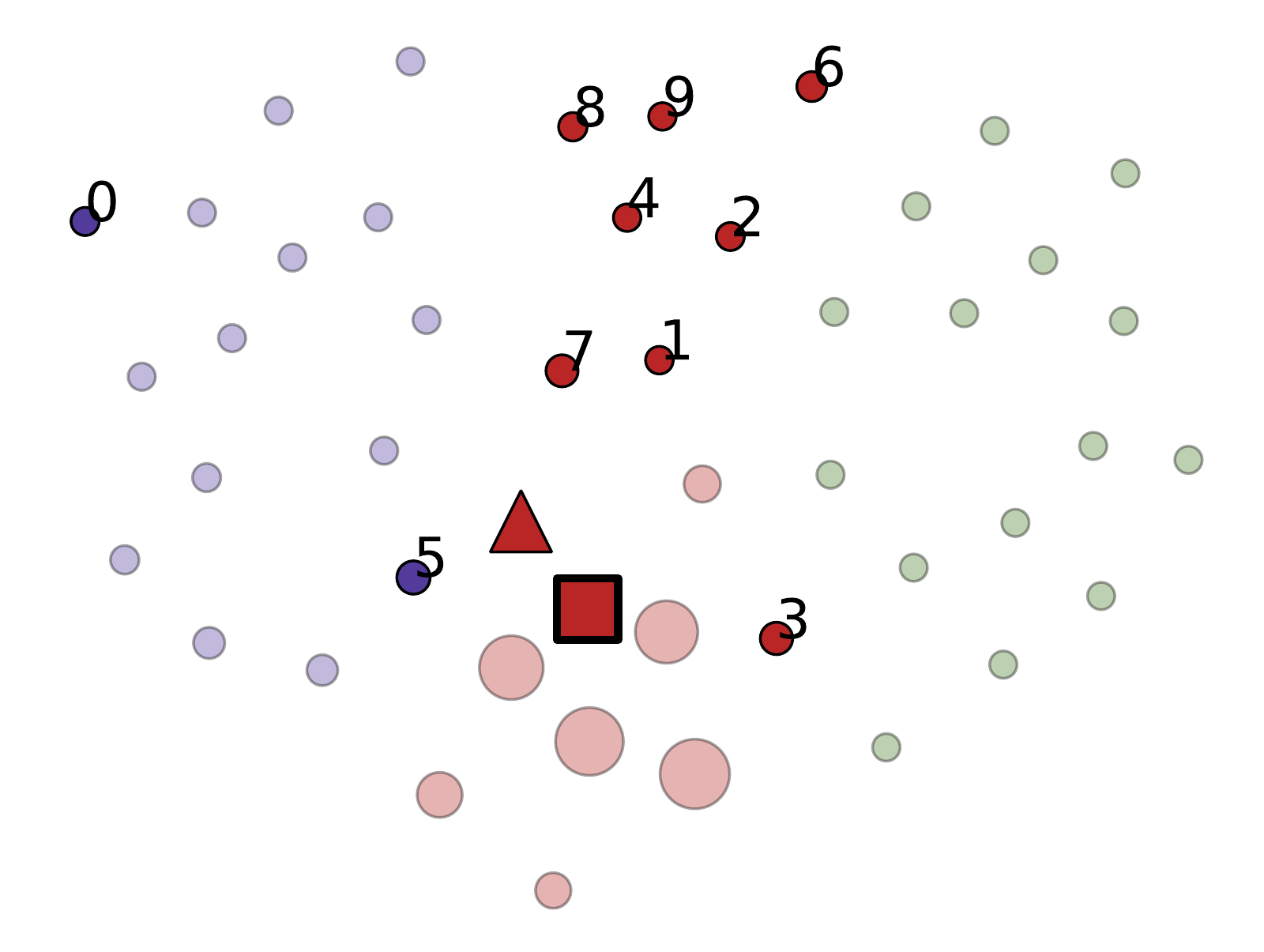}}
\hfil
    \subfigure[\small Flipped image, Full-path, Ensemble conf]{\includegraphics[width=0.42\linewidth]{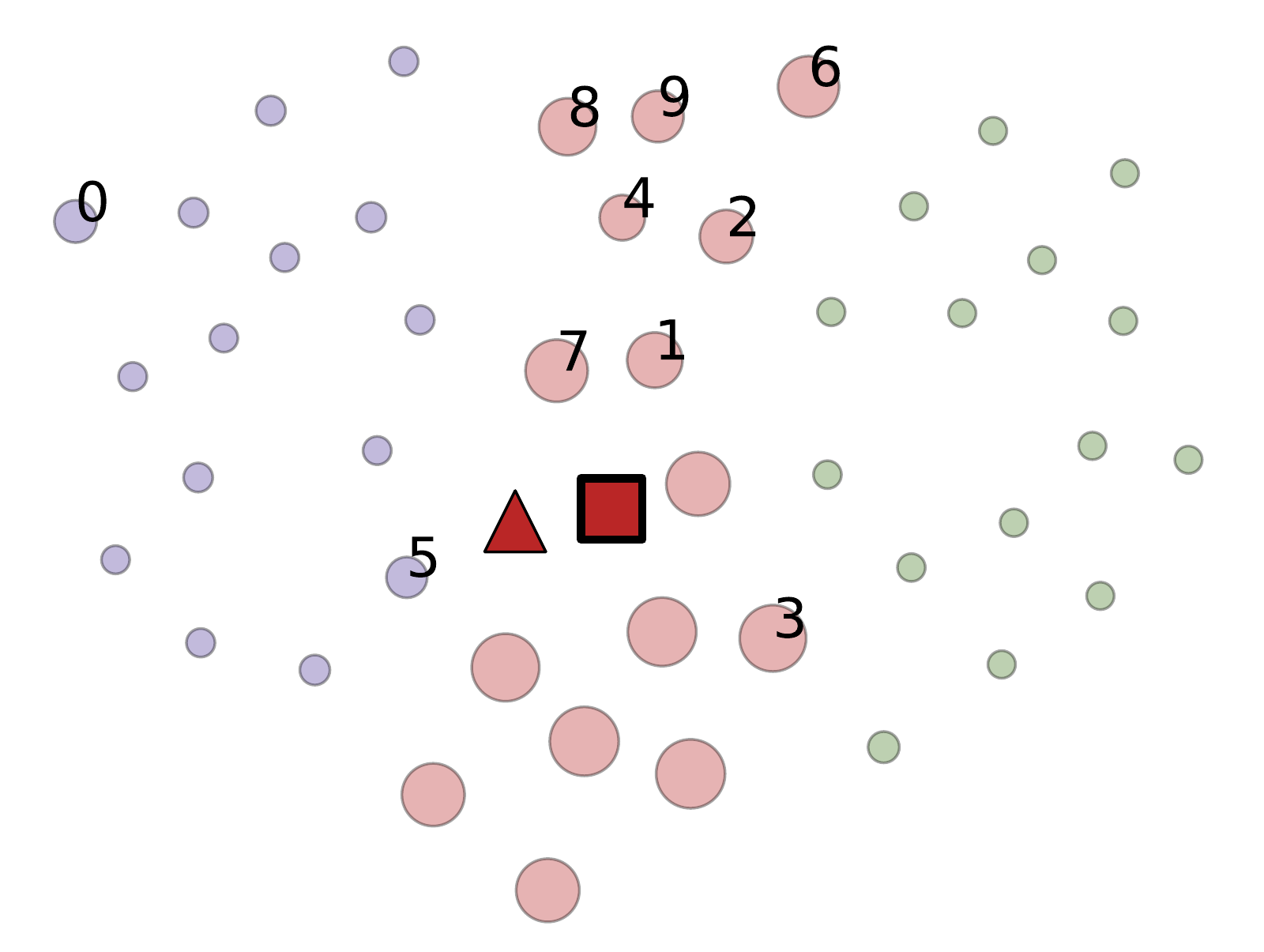}}
\vspace{-0.1in}
    \subfigure[\small Flipped image, Drop-path, Local conf]{\includegraphics[width=0.42\linewidth]{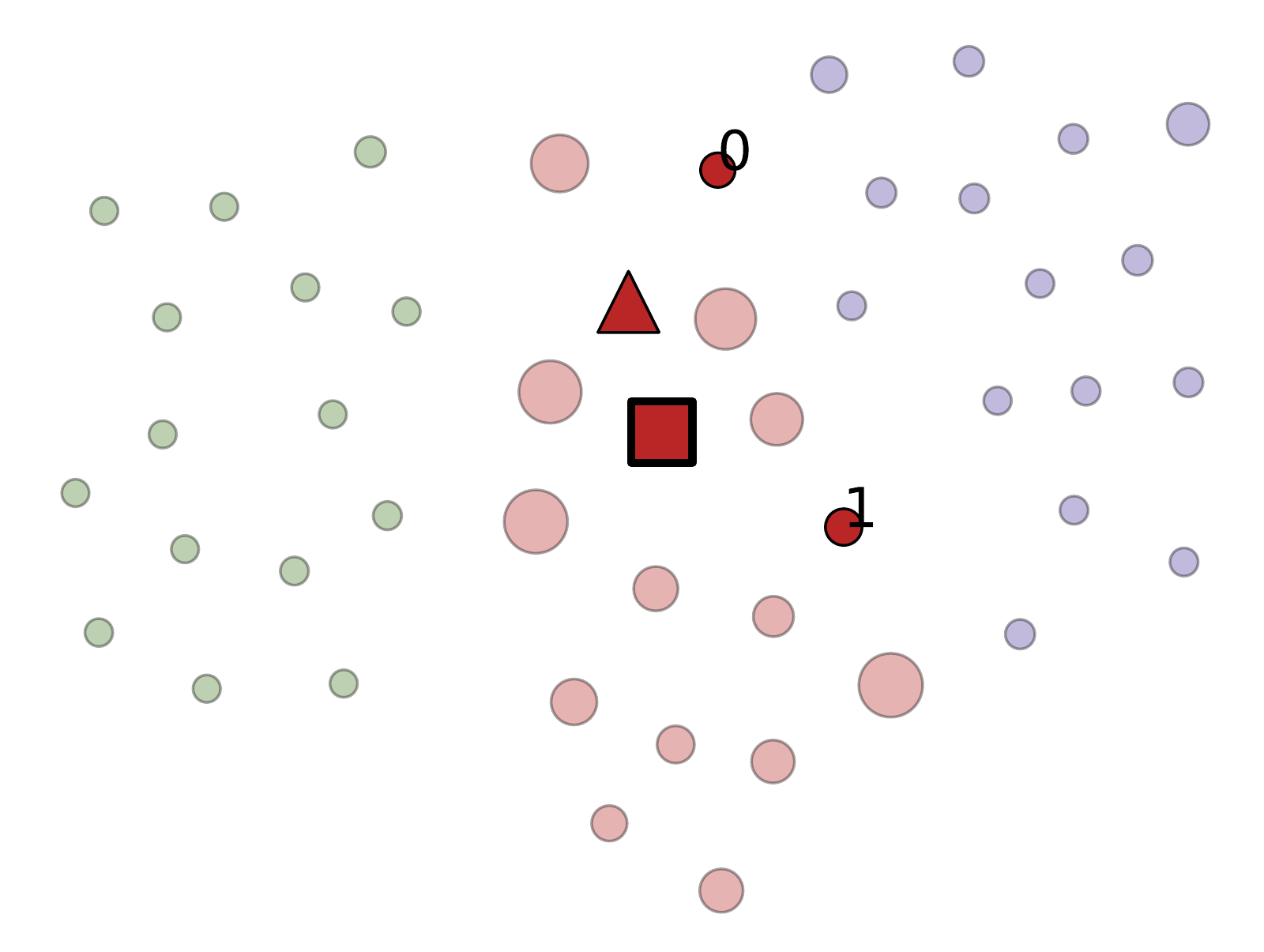}}
\hfil
    \subfigure[\small Flipped image, Drop-path, Ensemble conf]{\includegraphics[width=0.42\linewidth]{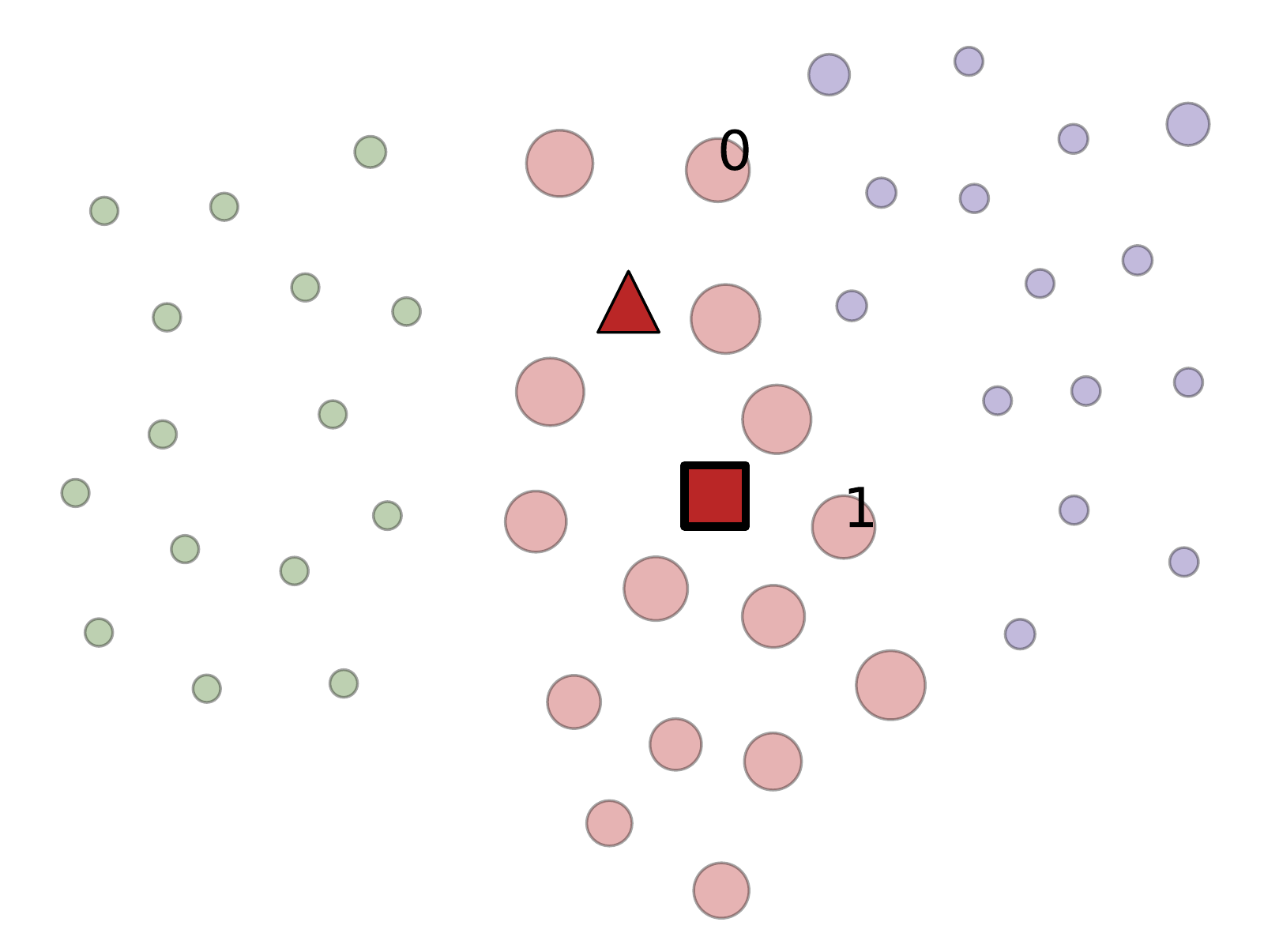}}
\caption{ \small Visualization of incorrectly classified query examples, on a miniImageNet 3-way 1-shot task. The size of circles shows the confidence score for the red class. Every figure is visualized by same task. conf denotes confidence. In each row, we show the transduction with local confidence and the transduction with ensemble confidence, where local confidence is derived from each space. Best viewed in color.}
    \label{fig:my figure}
    \end{figure}

% \begin{figure}[t!]
% 	\vspace{-0.1in}
% 	\centering
% 	\hfill
% 	\subfigure[ProtoNets (+4.44\%)]{\includegraphics[width=4.3cm]{pdfs/embedding_PN_4.44.pdf}\label{fig:fig1_PN}}
%     \hfill
% 	\subfigure[Instance-wise metric (+8.89\%) ]{\includegraphics[width=4.3cm]{pdfs/embedding_IMS_8.89.pdf}\label{fig:fig1_IMS}}
% 	\hfill
% 	\subfigure[MCT (+15.56\%)]{\includegraphics[width=4.3cm]{pdfs/embedding_MCT_15.56.pdf}\label{fig:fig1_MCT}}
% 	\hfill
% 	\vspace{-0.13in}
%     \caption{ \small Transductive inference with confidence scores. We visualize t-SNE embeddings on a 3-way 1-shot task, where each color stands for different class. The numbers show the accuracy increase after transduction for this task. The transparency shows the confidence scores for \emph{red} class.}
% 	\vspace{-0.20in}
%     \label{fig:concept_3way}
% \end{figure}
    \end{appendices}
    
\end{document}